\documentclass{article} % For LaTeX2e
\usepackage{iclr2019_conference,times}

\usepackage{amsmath,amsfonts,bm}

\def\eqref#1{equation~\ref{#1}}
\def\1{\bm{1}}

\def\eps{{\epsilon}}

\DeclareMathAlphabet{\mathsfit}{\encodingdefault}{\sfdefault}{m}{sl}
\SetMathAlphabet{\mathsfit}{bold}{\encodingdefault}{\sfdefault}{bx}{n}

\usepackage{hyperref}
\usepackage{url}
\usepackage{graphicx}
\usepackage{comment}
\usepackage[title]{appendix}

\title{Discriminative 
  out-of-distribution detection\\ 
  for semantic segmentation} 

\author{Petra Bevandi{\' c}, Sini{\v s}a {\v S}egvi{\' c}, Ivan Kre{\v s}o \& Marin Or{\v s}i{\' c} \\
Faculty of Electrical Engineering and Computing\\
Zagreb, Croatia \\
\texttt{\{name.surname\}@fer.hr}
}

\iclrfinalcopy % Uncomment for camera-ready version, but NOT for submission.
\begin{document}

\maketitle

\begin{abstract}
Most classification and segmentation datasets
assume a closed-world scenario in which 
predictions are expressed as distribution 
over a predetermined set of visual classes.
However, such assumption implies unavoidable 
and often unnoticeable failures
in presence of out-of-distribution (OOD) input.
%visual concepts that are foreign to the training distribution.
These failures are bound to happen 
in most real-life applications
since current visual ontologies 
are far from being comprehensive.
We propose to address this issue 
by discriminative detection 
of OOD pixels in input data.
Different from recent approaches,
we avoid to bring any decisions
by only observing the training dataset
of the primary model trained to solve 
the desired computer vision task.
% due to unavoidable epistemic uncertainty in presence of outliers.
Instead, we train a dedicated OOD model
which discriminates the primary training set
from a much larger "background" dataset 
which approximates the variety of the visual world.
We perform our experiments 
on high resolution natural images
in a dense prediction setup. 
We use several road driving datasets
as our training distribution,
while we approximate 
the background distribution with the ILSVRC dataset.
%TODO
We evaluate our approach on WildDash test,
which is currently the only public test dataset 
that includes out-of-distribution images.
The obtained results show 
that the proposed approach
succeeds to identify 
out-of-distribution pixels
while outperforming previous work 
by a wide margin.
\end{abstract}

\section{Introduction}

Development of deep convolutional models has resulted 
in tremendous advances of visual recognition.
%\cite{krizhevsky12nips,russakovsky15ijcv,he16eccv,huang17cvpr}.
Recent semantic segmentation systems 
surpass 80\% mIoU \citep{chen17arxiv}
on demanding natural datasets
such as Pascal VOC 2012 \citep{everingham15ijcv} 
or Cityscapes \citep{cordts15cvprw}.
Such performance level suggests 
a clear application potential
in exciting areas such as 
road safety assessment or autonomous driving.
Unfortunately, most existing semantic segmentation datasets 
assume closed-world evaluation \citep{scheirer13pami},
which means that they require predictions 
over a predetermined set of visual classes.
%are very useful for promoting research in the field up to a certain point.
Closed-world datasets are very useful for promoting research,
however they are poor proxies for real-life operation
even in a very restricted scenario such as road driving.
In fact, one can easily imagine many real-life driving scenes
which give rise to image regions 
that can not be recognized 
by learning on the Cityscapes ontology.
Some of those regions may be projected from objects 
which are foreign to Cityscapes 
(e.g.\ road works, water, animals).
Other may appear unrelated to Cityscapes 
due to particular configurations 
being absent from the training dataset
(e.g.\ pedestrians lying on the ground, crashed cars, fallen trees). 
Finally, some regions may be poorly classified 
due to different environmental conditions, acquisition setup, 
or geographical location \citep{tsai18cvpr}.
This indicates that our models rely 
on factors of variation which are dataset specific 
(e.g. time of day, weather, particular camera etc.)
rather than on the gist of the visual stimulus.

The simplest way to approach unrecognizable data
is to improve datasets.
For instance, the Vistas dataset \citep{neuhold17iccv}
proposes a richer ontology and addresses 
more factors of variation than Cityscapes.
However, training on Vistas requires
considerable computational resources
while still being unable to account for the full variety 
of the recent WildDash dataset \citep{Zendel_2018_ECCV},
as we show in experiments.
Another way to approach this problem 
would be to design strategies for knowledge transfer
between the training dataset and 
the test images \citep{tsai18cvpr}.
However, this is unsatisfactory
for many real world applications 
where the same model should be directly applicable
to a variety of environments.

These examples emphasize the need to quantify 
model prediction uncertainty \citep{kendall17nips},
especially if we wish to achieve 
reliable deployment in the real world.
%since the ability to detect 
%incomprehensible events is crucial 
%for reliable deployment in the real world.
Unfortunately, this is not easy to achieve, 
since uncertainty occurs in two distinct flavours.
Aleatoric uncertainty \citep{kendall17nips} 
is caused by inherent volatility of the data.
For example, small transport vehicles 
are sometimes labeled as class car 
while elsewhere they may be labeled as class truck;
semantic borders can not be predicted correctly
since most models perform inference at 4$\times$
or even more subsampled resolution.
However, these issues could be resolved 
by improving the annotation standard, and
excluding semantic borders from the evaluation 
\citep{everingham15ijcv}.
%or hierarchical softmax evaluation.
We therefore believe that aleatoric uncertainty
is merely a consequence of imperfect
annotation and evaluation protocols,
rather than an inherent obstacle to machine intelligence.
On the other hand, 
epistemic uncertainty \citep{kendall17nips}
arises when the trained model
is unable to bring the desired prediction
given particular training dataset.
In other words, it occurs when the model 
receives the kind of data 
which was not seen during training.
Epistemic uncertainty is therefore strongly related 
to the probability that the model operates
on an out-of-distribution sample.

Several researchers have attempted to detect
out-of-distribution samples
by recognizing high entropy in the model output
\citep{kendall17nips,hendrycks17iclr,liang18iclr,devries18arxiv}.
Unfortunately, such approaches are unable 
to decouple epistemic from aleatoric uncertainty
and therefore achieve poor precision 
in our semantic segmentation experiments.

Therefore, in this paper we propose a novel approach
which detects out-of-distribution samples 
by a dedicated "OOD" model being completely separate from 
the "primary" model trained for a specific vision task.
The only responsibility of the OOD model is to predict 
a probability that the sample $\mathbf{x}$ is an outlier:
$P(\mathrm{outlier}|\mathbf{x},\theta)$.
We formulate the OOD model 
as binary classification 
between the training dataset 
and a much larger "background" dataset.
Ideally, the background dataset would contain
the entire diversity of the visual world,
however, our experiments show that 
ILSVRC \citep{russakovsky15ijcv}
can be used as a suitable approximation
\citep{oquab14cvpr,gasulla18jair}.
The proposed formulation is insensitive 
to any kind of aleatoric uncertainty 
related to the particular computer vision task.
This distinct advantage allows the proposed approach
to deliver significantly better performance 
than related approaches in which OOD detection
is coupled with the primary model.

\begin{comment}
The rest of this paper is organized as follows.
Section 2 presents a review of the related work
in the field of assessing the prediction uncertainty.
Section 3 presents adaptations of several previous
prediction uncertainty approaches 
for the semantic segmentation task.
The proposed approach for recognizing 
out-of-distribution samples as ImageNet content
is detailed in Section 4.
Section 5 presents and discusses experimental results
while the conclusions are presented in Section 6.
\end{comment}

\section{Related work}

Detection of out-of-distribution (OOD) examples 
has received a lot of attention in recent literature.
Many approaches try to estimate uncertainty
by analyzing entropy of the predictions.
The simplest approach is to express the prediction confidence
as the probability of the winning class or, equivalently, 
the maximum softmax (max-softmax) activation \citep{hendrycks17iclr}.
The resulting approach achieves useful results
in image classification context,
although max-softmax must be recalibrated \citep{guo17icml}
before being interpreted as $P(\mathrm{inlier}|\mathbf{x})$.
This result has been improved upon 
by the approach known as ODIN \citep{liang18iclr}
which proposes to pre-process input images 
with a well-tempered anti-adversarial perturbation 
with respect to the winning class 
and increase the softmax temperature.

%many patches in semantic segmentation datasets 
%contain more than one semantic class,
%while such patches are very rare in classification datasets.
Unfortunately, these approaches do not distinguish 
between aleatoric and epistemic uncertainty.
Consequently, they are not easily ported to dense prediction
where there is a high incidence of aleatoric uncertainty.
Consider the output of a semantic segmentation model
at an inlier pixel located on a semantic border:
the prediction is likely going to be highly uncertain 
even when the corresponding patch is not an outlier.
Other sources of aleatoric uncertainty include 
distant and underexposed parts of the scene
where it is very hard to reach the right decision,
as well as ambiguous semantic labels 
(e.g. class fence vs class wall vs class building, 
      class road vs class low sidewalk etc).
Most of such situations lead to false positive OOD responses
and severely reduce AP performance.

%Other approaches have tried to detect anomalies 
%at the feature level \citep{mandelbaum17arxiv},
%and to train the classifier to emit high entropy prediction
%for artificial samples generated by a GAN \citep{lee18iclr}.
Another line of research characterizes uncertainty
by a separate head of the primary model 
which learns either prediction uncertainty 
\citep{kendall17nips,lakshminarayanan17nips} 
or confidence \citep{devries18arxiv}.
The predicted variance (or confidence) 
is further employed to discount the data loss
while at the same time incurring a small penalty 
proportional to the uncertainty 
(or inversely proportional to confidence).
This way, the model is encouraged to learn to recognize hard examples
if such examples are present in the training dataset.
Unfortunately, such modelling is able to detect 
only aleatoric uncertainty \citep{kendall17nips},
since the data which would allow us
to learn epistemic uncertainty is absent by definition.

A principled information-theoretic approach 
for detecting out-of-distribution samples
in presence of aleatoric uncertainty
has been proposed by \citet{smith18uai}.
They express epistemic uncertainty as 
mutual information between the model parameters 
and the particular prediction.
Intuitively, if our knowledge about the parameters
increases a lot when the ground truth prediction becomes known,
then the corresponding sample is likely to be out of distribution.
The sought mutual information is quantified as 
a difference between the total prediction entropy 
and the marginalized prediction entropy 
over the parameter distribution.
Both entropies are easily calculated with MC dropout.
Unfortunately, our experiments along these lines 
resulted in poor OOD detection accuracy.
%due to estimated epistemic uncertainty 
%being affected by aleatoric uncertainty.
%which indicates that the two uncertainties 
%were not successfully decoupled.
This may be caused by MC dropout being 
an insufficiently accurate approximation 
of the model sampling according to the parameter distribution,
however, further work would be required
in order to produce a more definitive answer.

\citet{lakshminarayanan17nips} 
show that uncertainty can be more accurately recovered
by replacing MC dropout with an ensemble 
of several independently trained  models.
However, for this to be done,  
many models need to be in GPU memory during evaluation. 
Explicit ensembles are therefore not suited for systems 
which have ambition to perform
dense prediction in real time.

Out-of-distribution samples can also be detected 
by joint training on the primary dataset and some "foreign" dataset,
e.g. Cityscapes (road driving) and ScanNet (indoor).
Out-of-distribution sample is signaled
whenever the prediction favours a class 
which is foreign to the evaluation dataset \citep{kreso18arxiv}.
For instance, if a bathtub region is detected 
in an image acquired from an intelligent vehicle, 
then the pixels of that region are likely OOD.   
%This work also points out the importance of mixed batches 
%when jointly training on multiple training datasets.

Contrary to all previous OOD detection approaches,
we attempt to cast the problem as supervised discrimination
between the primary training dataset and the "background" distribution
which we approximate with the ILSVRC dataset.
In the proposed formulation, the OOD model 
is trained on much more data than the primary model.
This is not an easy path, since it requires 
dealing with inevitable contention 
between the primary training dataset and ILSVRC images.
Nevertheless, such approach has none of the serious problems
inherent to the approaches presented above,
and so it may be the most feasible solution to the problem at hand.

\section{The proposed discriminative OOD detection approach}

%Following the difficulties of existing models
%to recognize out of distribution data,
%we try to understand why is that task so difficult.
A principled algorithm to recognize OOD samples
would fit a generative model $P_\mathcal{D}(\mathbf{x}|\theta)$
to the training dataset $\mathcal{D}=\{\mathbf{x}_i\}$.
Such model would learn to evaluate the probability distribution 
of the training dataset at the given sample.
%In other words, the model outputs the probability 
%that the given sample is drawn from $\mathcal{D}$ by random sampling.
Clearly, the output of such a model would be closely related 
to epistemic uncertainty
of the model prediction in $\mathbf{x}$.
Unfortunately, this approach would be easily implemented
only for low-dimensional samples and not for images.
The most successful generative model for images -- GAN
\citep{goodfellow14nips} --
is unable to evaluate the probability distribution 
and instead focuses on generating images from random noise.
One could try to address the problem more directly,
by attempting to detect outliers with a GAN discriminator.
% TODO: cite članak po kojem je Josip radio diplomski?
However, this would result in random predictions
(at least this was the case in our experiments)
since the discriminator is not trained to recognize 
all possible kinds of outliers as fake samples.
% spomenuti Šarićev diplomski?

Nevertheless, the idea to avoid modeling 
the inlier distribution (which is hard)
and instead simply learning to detect outliers
through supervised loss sounds very interesting.
The key question at this point is:
on which data should the desired
binary classifier be learned.
Clearly, the in-distribution class should correspond to 
the training dataset of the primary model (e.g. Cityscapes train).
On the other hand, the out-of-distribution class 
should contain patches from all possible natural images, 
or, in other words, entire diversity of the visual world.
Since none of existing public datasets
is able to satisfy our requirements,
in this paper we propose to use 
the ILSVRC dataset \citep{russakovsky15ijcv}
as the next-best solution for this purpose.
We think that this approximation is reasonable
since the ILSVRC dataset has been successfully used 
in many knowledge transfer experiments 
\citep{oquab14cvpr,gasulla18jair}.
The ILSVRC dataset is also interesting since 
approximately half of its images 
include the bounding box 
of the object which defines the class.
This information can be exploited to alleviate 
the problems which we discuss next.

Now that we have settled the training data,
we need to define learning algorithm in a way
to avoid ILSVRC images hijacking content 
from the primary training dataset.
We observe two distinct problems in that context:
i) ILSVRC classes which overlap classes from the primary dataset,
   e.g.\ a car in an ILSVRC image labeled as car,
and ii) primary dataset classes in the ILSVRC image background,
   e.g.\ a person playing a banjo in an ILSVRC image labeled as banjo.
Currently, we address both problems by ensuring 
that the model is trained on ID pixels 
as often as on OOD ones. 
Due to diversity of the OOD class (ILSVRC)
such training results in a bias 
towards the ID dataset. 
For example, there are around 10 car-like classes in ILSVRC; 
therefore, cars occur in only around 1/100 ILSVRC images. 
On the other hand, almost every image 
from any road driving dataset 
is likely to contain at least one car. 
Hence, the proposed approach ensures that
the model is more likely to classify 
car pixels as inliers (Cityscapes) 
rather than as outliers (ILSVRC).
%TODO
We consider several improvements to this approach in the appendix.

% traffic classes in different classes 
% (e.g. a man in traffic vs a man playing an instrument)
%We address the second problem by exploiting the fact 
%that most ImageNet images are very well aligned: 
%the defining object is almost always located in the image center.
The proposed OOD model is discriminative and fully convolutional, 
while we train it with the usual cross-entropy loss.
During evaluation, we apply the trained model
to test images in order to obtain 
dense logit maps for the two classes.
These logits reflect the epistemic uncertainty
in each individual pixel.
OOD detection is finally performed 
by thresholding the probability 
%or the logit 
of the inlier class.

% premjestiti u 5:
% cityscapes vs vistas vs wd as ID examples (cityscapes too specific)

\section{Experiments}
\label{sec:exp}

Our experiments explore the accuracy of dense OOD detection 
in realistic natural images from several driving datasets.
We train our models on different datasets
and discuss the obtained performance 
on WildDash test and other datasets.
Our experiments compare the proposed approach (cf.\ Section 3)
to several previous approaches from the literature,
which we adapt for dense prediction as explained next.
%Next we describe our experimental setup. 
%We begin by describing the WildDash dataset 
%which provides an immediate need for detecting OOD pixels. 
%Then we describe the semantic segmentation models 
%that were used and how they were trained. 
%Finally we describe and discuss our results.

\subsection{Adapting image-wide approaches for dense prediction}

We consider three previous approaches
which were originally developed 
for image-wide OOD detection
and adapt them for dense prediction.
For the max-softmax approach \citep{hendrycks17iclr},
the task is trivial: it is enough to independently
assess the prediction of the primary model 
in each image pixel.

For the ODIN approach \citep{liang18iclr},
we first perturb the input image
in the direction which increases 
the probability of the winning class 
in each individual pixel 
(according to the primary model).
Consequently, we apply the softmax temperature
and consider the max-softmax response as above.
The perturbation magnitude $\eps$ and softmax temperature $T$
are hyperparameters that need to be validated.
%: $\eps$=$5\cdot10^{-4}$, T=100.
Note that dense ODIN perturbation
implies complex pixel interactions
which are absent in the image-wide case.
This approach achieved a modest improvement
over the max-softmax approach
so we do not present it in the tables.

For trained confidence \citep{devries18arxiv}, 
we introduce a separate convolutional head to the primary model.
The resulting confidence predictions 
diminish the loss of wrong prediction in the corresponding pixel
while incurring a direct loss multiplied with a small constant.

\subsection{Datasets}

Cityscapes is a widely used dataset \citep{cordts15cvprw}
containing images from the driver perspective
acquired during rides through different German cities. 
It is divided into training, validation and test subsets. 
The training set contains 2975 annotated images, 
while the validation set contains 500 annotated images.

Vistas \citep{neuhold17iccv} is larger and more diverse than Cityscapes.
It contains much more diversity with respect to 
locations, time of day, weather, and cameras. 
There are 18\,000 train and 2\,000 validation images. 

The WildDash dataset \citep{Zendel_2018_ECCV} provides a benchmark
for semantic segmentation and instance segmentation. 
It focuses on providing performance-decreasing images.
These images are challenging due to conditions and 
unusual locations in which they were taken or because
they contain various distortions. 
%The images are divided into a validation set and a test set.
There are 70 validation and 156 test images.
The test set contains 15 images which are marked as negatives. 
All pixels in these images are considered out-of-distribution
in the context of semantic segmentation on road driving datasets.
These images contain noise, indoor images, and 
five artificially altered inlier images 
(see Figure \ref{fig:ID-as-OOD}). 
WildDash is compatible with Cityscapes labeling policy \citep{cordts15cvprw}, 
but it also considers performance on negative images. 
Pixels in negative images are considered to be correctly classified 
if they are assigned a correct Cityscapes label 
(e.g. people in an indoor scene) or if they are assigned a void label                                                
(which means that they are detected as OOD samples). 
The official WildDash web site suggest that OOD pixels 
could be detected by thresholding the max-softmax value.

The ILSVRC dataset \citep{russakovsky15ijcv} includes 
a selection of 1000 ImageNet \citep{deng09cvpr} classes.
It contains 1\,281\,167 images with image-wide annotations 
and 544\,546 images annotated with the bounding box
of the object which defines the class.
The training split contains over one million images, 
while the validation split contains 50\,000 images.

The Pascal VOC 2007 dataset contains 9\,963 training 
and validation images with image-wide annotations into 20 classes.
Pixel-level semantic segmentation groundtruth is available
for 632 images.

%TODO životinje

%TODO experiments on pasted animals and vistas animals

\subsection{Model details}
\label{sec:models}

We experiment with three different models:
i) the primary model for semantic segmentation,
ii) the augmented primary model with confidence 
    \citep{devries18arxiv},
and iii) the proposed discriminative model.
All three models are based on 
the DenseNet-121 architecture \citep{huang17cvpr} 
(we assume the BC variant throughout the paper).
DenseNet-121 contains 120 convolutional layers
which can be considered as a feature extractor. 
These layers are organized into 4 dense blocks (DB\_1 to DB\_4)
and 3 transition layers (T\_1 to T\_3) inbetween. 
In all of our models, the feature extractor is initialized 
with parameters pretrained on ILSVRC.

We build our primary model by concatenating 
the upsampled output of DB\_4 with the output of DB\_2. 
This concatenation is routed to the 
spatial pyramid pooling layer (SPP) \citep{he14eccv}
which is followed by a BN-ReLU-Conv block 
(batch normalization, ReLU activation, 3x3 convolution)
% SS: bilinear upsampling or downsampling the groundtruth? 
which outputs 19 feature maps with logits.
The logits are normalized with softmax and 
then fed to the usual cross-entropy loss. 
%We jointly train the feature extractor (DB\_1 to DB\_4)
%and the semantic segmentation head, 
%however we update the feature extractor parameters
%with a smaller learning rate. 

We augment the primary model by introducing 
a new SPP layer and BN-ReLU-Conv block parallel to 
the SPP layer and BN-ReLU-Conv layer of the 
primary model.
The output of confidence BN-ReLU-Conv block is
concatenated with the segmentation maps.
The confidence estimation is obtained 
by blending the resulting representation 
with a BN-ReLU-Conv block with sigmoid activation. 
We prevent the gradients from flowing 
into the segmentation maps 
to ensure that the segmentation head 
is only trained for segmentation 
and not for estimating confidence. 
The confidence map is then used to modulate 
the cross-entropy loss of the segmentation maps
while low confidence is penalized 
with a hyper-parameter $\lambda$. 
%The two heads and the feature extractor are jointly trained, 
%but we update the feature extractor parameters with a smaller learning rate.

Our proposed discriminative OOD detector 
feeds the output of DB\_4 to a BN-ReLU-Conv block 
with 2 feature maps corresponding to logits 
for inliers and outliers. 
The logits are fed to softmax
and then to the cross-entropy loss
which encourages the model to classify 
the pixels according to the respective labels.  
We only train the head, DB\_4 and T\_3 in order to 
speed up learning and prevent overfitting.
We assume that initial DB\_3 features 
are expressive enough due to ILSVRC pretraining, 
and that it is enough to only fine-tune DB\_4 
for discriminating road-driving scenes 
from the rest of the visual world. 
%The learning rate for DB\_4 is four times smaller 
%than the learning rate for BN-ReLU-Conv block.

\subsection{Training}
\label{sec:train}

We train the primary model on Cityscapes train at half resolution.
The training images are normalized with 
Cityscapes train mean and variance.
We optimize the model on entire images for 40 epochs 
without jittering using ADAM optimizer, 
a decaying learning rate starting from 4e-4 and a batch size of 5.
The learning rate for the pretrained DenseNet layers is four times smaller
than the learning rate for the model head.
The model achieves 70.1 \% mIoU on Cityscapes val 
and 23.6 \% mIoU on WildDash val.
Due to poor generalization on WildDash, 
we also trained a model on the combination 
of Cityscapes train and WildDash val. 
This model reaches 70.2\% mIoU 
on the Cityscapes validation set.

We train the augmented primary model \citep{devries18arxiv} 
in the same way as the primary model. 
This model achieves 71.1 \% mIoU on Cityscapes val 
and 25.7 \% mIoU on WildDash val.

We train our discriminative models for OOD detection
in the similar way as the primary model. 
We use three different in-distribution (ID) datasets: 
Cityscapes, Cityscapes + WildDash val and Vistas.
WildDash val was added to Cityscapes because 
models trained on Cityscapes are prone to overfitting.
In order to show that training 
on the WildDash validation subset can be avoided,
we also show results of a model instance 
trained on the Vistas dataset \citep{neuhold17iccv}
which is much more diverse than Cityscapes. 

As explained earlier, we always use ILSVRC 
as the training dataset for the OOD class.
Unfortunately, ILSVRC images
are not annotated at the pixel level. 
We deal with this challenge 
by simply labeling all ILSVRC pixels as the OOD class,
and all pixels from the road-driving datasets as the ID class.
% TODO
%In later experiments we repeat the same procedure,
%however we use only the image part which is 
%inside the annotated groundtruth bounding box.
%The latter approach can use only around 50\% 
%ILSVRC images for which bounding box is available.

In order to account for different resolutions, 
we resize the ILSVRC images 
so that the smaller side equals 512 pixels 
while keeping the image proportions. 
We form mixed batches \cite{kreso18arxiv}
by taking random crops of 512$\times$512 pixels
from both training datasets (ILSVRC, in-distribution)
and normalizing them using the ILSVRC mean and variance. 
Since there is a huge disproportion in size 
between the ILSVRC dataset and road-driving datasets, 
we oversample the road-driving datasets 
so that the number of images becomes approximately equal.
We train the models using ADAM optimizer, 
a decaying learning rate and a batch size of 30,
until the accuracy on WildDash val reaches 100\%.
Similar accuracies are also observed 
on ILSVRC val (OOD) and Vistas val (ID)
after the training.

\subsection{Evaluation}

%SS: ovo isto kaže i naredni odlomak pa sam ukomentirao
%To evaluate how well the trained models separate OOD and ID, 
%we calculate average precision of OOD detection on WildDash benchmark set. 
%Additionally, we were provided the model of the runner-up to the ROB Challenge 
%for comparison with the models that we have trained. 
% For this model we consider the sum of probabilites 
%over softmax values for cityscapes classes to represent the probability that the pixel is ID.

We evaluate how well the considered models 
separate OOD and ID samples 
on several test datasets. 
% SS: ovo kažemo dolje, ne treba se ponavljati
% we calculate average precision of OOD detection on WildDash test. 
We compare our discriminative OOD model 
with the max-softmax \citep{hendrycks17iclr},
ODIN \cite{liang18iclr}, trained confidence \citep{devries18arxiv}
and the pretrained runner-up model from the ROB Challenge
which was provided by its authors \citep{kreso18arxiv}.
We quantify performance with average precision 
because it shows how well the method 
separates the OOD and ID pixels 
without having to look for appropriate discrimination thresholds.

%TODO
%In our first experiment 
We assume image-wide ID and OOD labeling
(further experiments are presented in the appendix).  % TODO
We label all pixels in WildDash test images \#0-\#140 as ID, 
and all pixels in WildDash test images \#141-\#155 as OOD. 
We provide two AP measures. 
The first one evaluates results on all negative images (\#141--\#155).
The second one ignores altered valid images \citep{Zendel_2018_ECCV}
(see Figure \ref{fig:ID-as-OOD}) which, in our view, 
can not be clearly distinguished from in-distribution images.
Thus, we respect the original setup of the challenge, 
but also provide the second performance metric 
for a more complete illustration of the results.
%because the WildDash benchmark 
%considers Cityscapes segmentations of those images 
%as valid (meaning it accepts both answers). 
%We do not go into more complex details. 
%That means that our test procedure considers 
%pixels at Cityscapes classes in negative images as OOD (e.g. people in a room), 
%while positive image pixels which do not belong to Cityscapes classes 
%are treated as ID (e.g. animals on the roads).
Note that we consider all pixels in OOD images as "positive" responses,
including the pixels of Cityscapes classes 
(e.g.\ a person in a room).
Conversely, we consider all pixels in ID images as "negatives",
including the pixels which belong to some of the 
ILSVRC classes (e.g. animals on the road). 
Such ambiguous pixels are rare and do not compromise our conclusions.

%TODO experiments on pasted animals and vistas animlas

\subsection{Results on WildDash test}

\begin{figure}[htb]
  \centering
  \includegraphics[width=1\columnwidth]{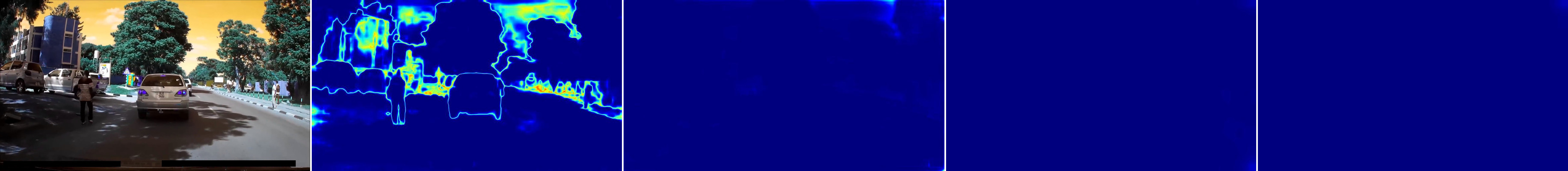}
  \includegraphics[width=1\columnwidth]{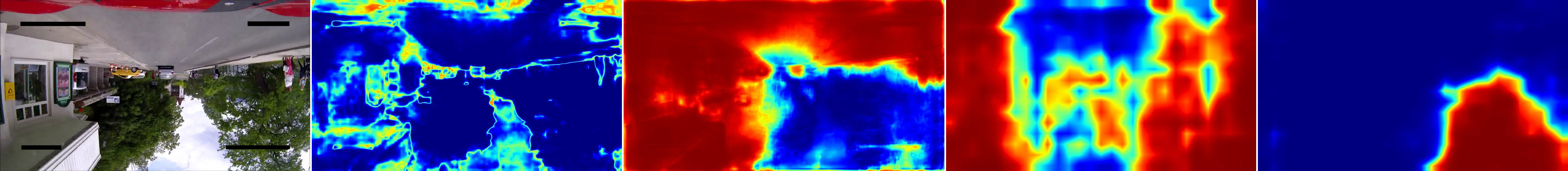}
  \includegraphics[width=1\columnwidth]{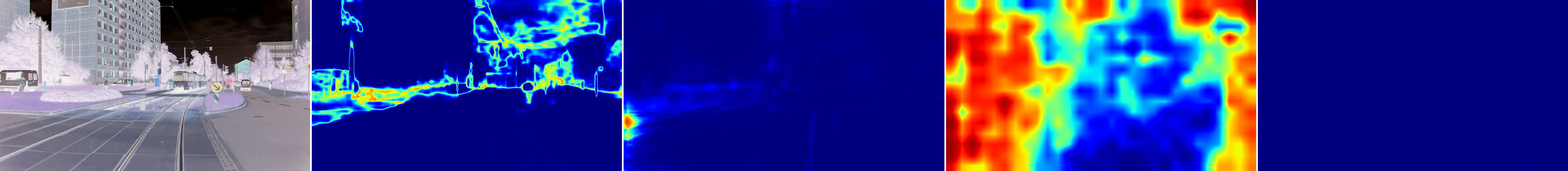}
  \caption{OOD detection in three of the five 
    altered valid scenes from WildDash test 
    (images \#141, \#148 and \#151).
    These images were ignored in experiments 
    labeled as "WD test selected". 
    The columns correspond to:
    i) original image, 
    ii) best result from Table \ref{table:APseg}, 
    iii) OOD detection with the ROB model, 
    iv) discriminative OOD detection trained on Vistas, 
    and v) discriminative OOD detection 
    trained on Cityscapes and WildDash val.
    Red denotes the confidence 
    that the corresponding pixel is OOD
    which can be interpreted as epistemic uncertainty.
  }
  \label{fig:ID-as-OOD}
\end{figure}

Table \ref{table:APseg} shows the results 
of OOD detection based on the max-softmax criterion
with two instances of our primary semantic segmentation model.
The two model instances were trained on i) Cityscapes train and 
ii) Cityscapes train + WildDash val.
The evaluation was performed on the WildDash test 
without the five altered valid images 
(cf.\ Figure \ref{fig:ID-as-OOD}) (left)
and on the complete WildDash test (right). 
Both approaches perform rather poorly, though
training on WildDash val doubles the performance.

\setlength{\tabcolsep}{4pt}
\begin{table}[htb]
\begin{center}
\caption{Average precision performance for OOD detection 
  by applying max-softmax to predictions
  of the primary model trained on two road-driving datasets.}
\label{table:APseg}
\begin{tabular}{|c||c|c|}
  \hline
  Training set & 
    \multicolumn{1}{c|}{WD test selection} & 
    \multicolumn{1}{c|}{WD test complete} 
 \\
 \hline
 \hline
  \multicolumn{1}{|l||}{City}          & 10.09 & 11.91\\
\hline
  \multicolumn{1}{|l||}{City + WD val} & 17.62 & 19.29\\
\hline

\end{tabular}
\end{center}
\end{table}
\setlength{\tabcolsep}{1.4pt}

\begin{figure}[htb]
  \centering
  
  \includegraphics[width=0.244\columnwidth]{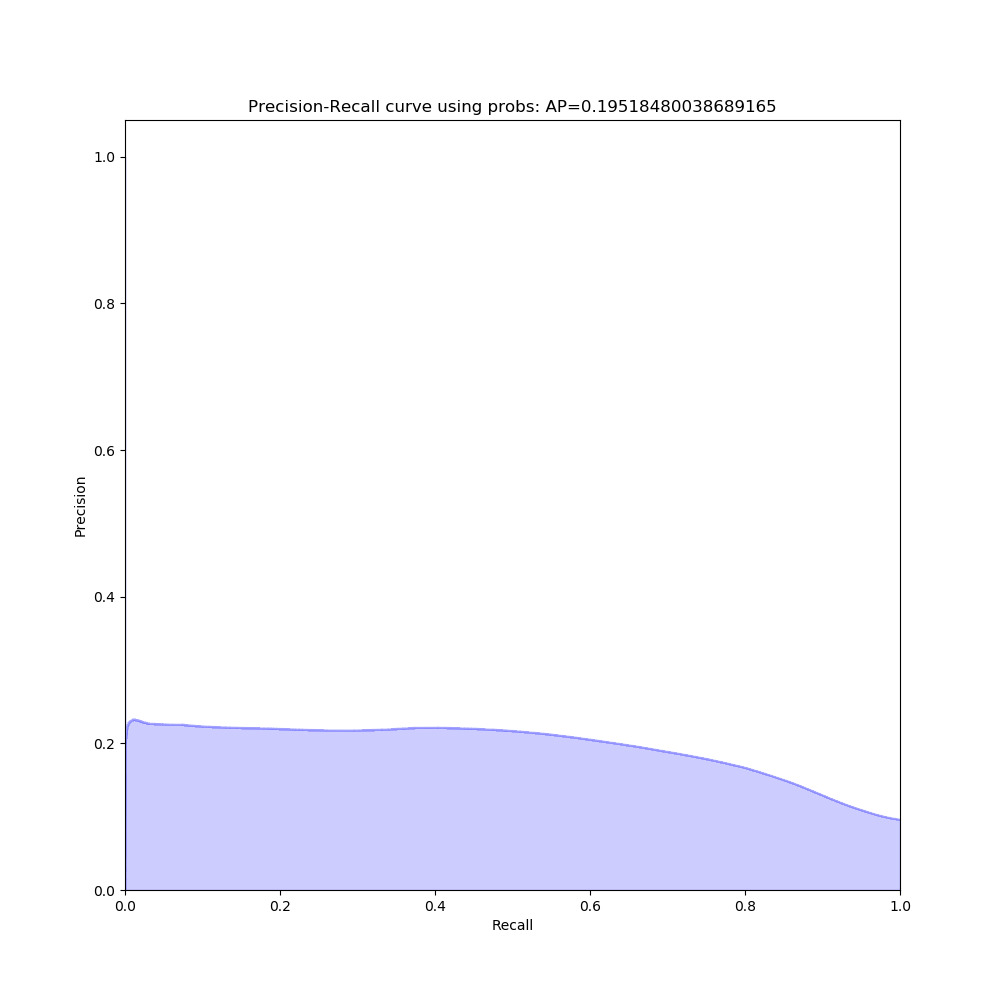}
  \includegraphics[width=0.244\columnwidth]{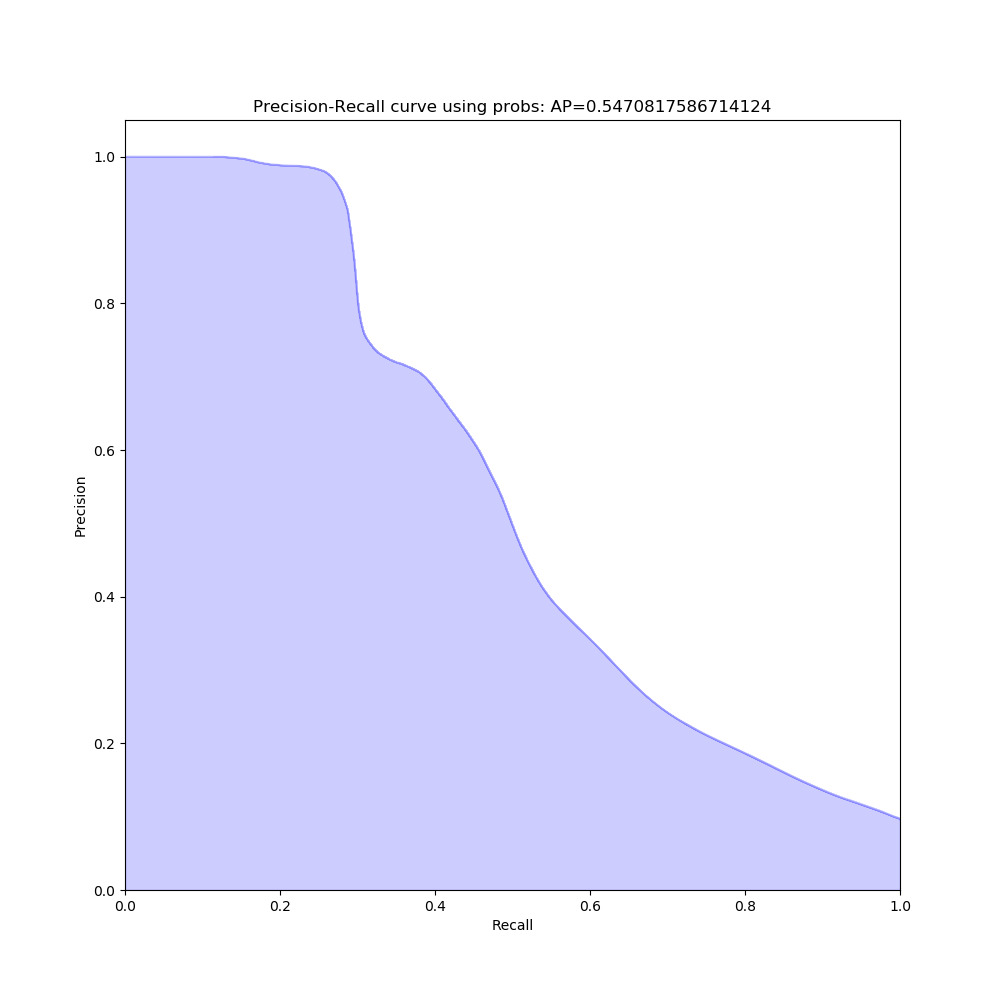}
  \includegraphics[width=0.244\columnwidth]{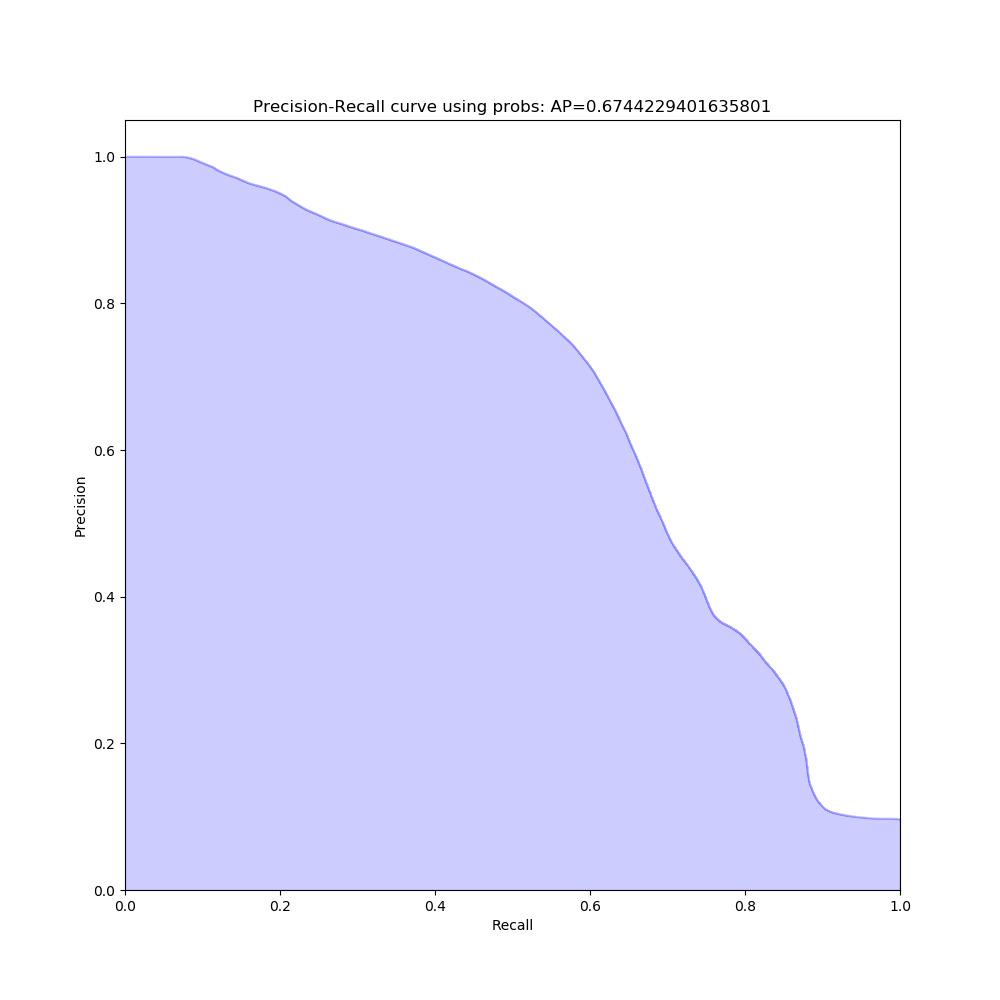}
  \includegraphics[width=0.244\columnwidth]{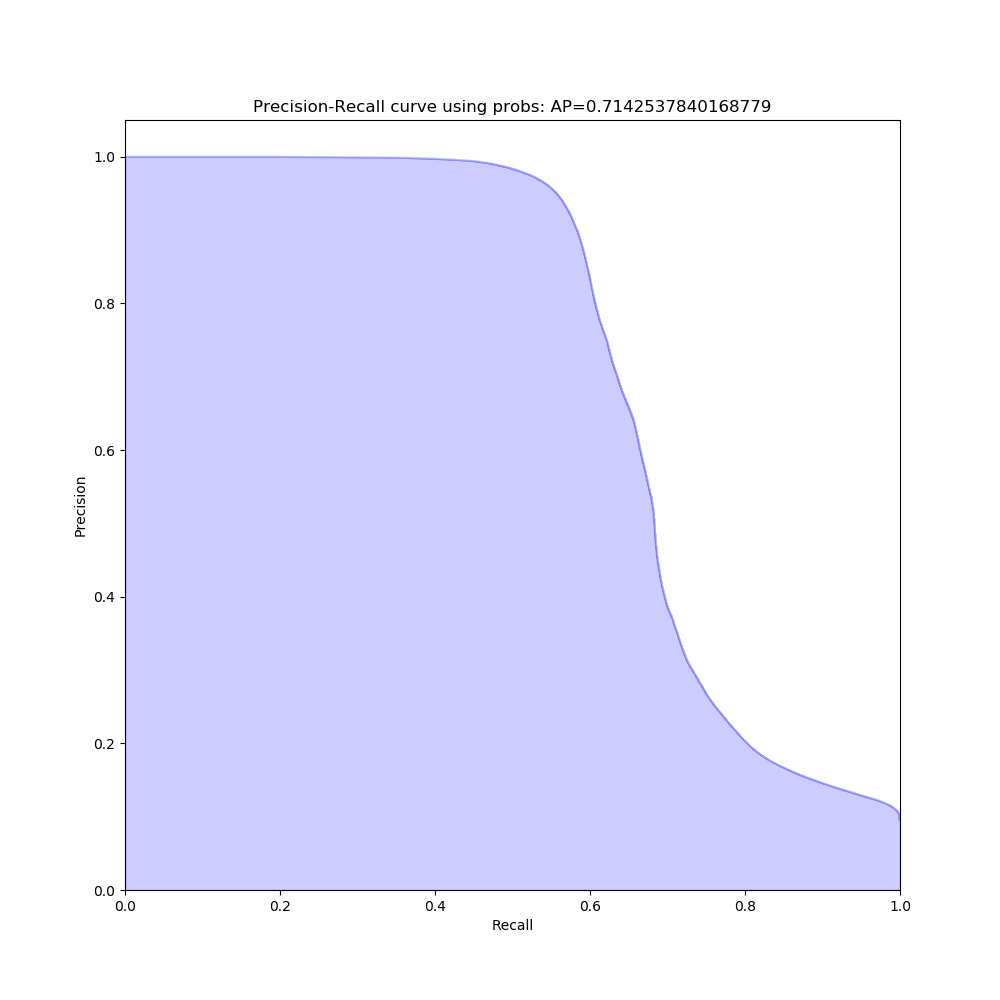}
  %, trim=2cm 2.5cm 2cm 16.8cm, clip
  \caption{Precision-recall curves of max-softmax OOD responses 
  	on the complete WildDash test. 
    The columns correspond to: 
    i) OOD detection according to max-softmax of the primary model, 
    ii) OOD detection by recognizing foreign classes 
        with the ROB model, 
    iii) discriminative OOD detection trained on Vistas, 
    and iv) discriminative OOD detection 
    trained on Cityscapes and WildDash val.
  }
  \label{fig:hists}
\end{figure}

The precision-recall curve for the model trained on Cityscapes and WildDash val can be seen in the first column of Figure \ref{fig:hists}. The precision is low even for small values of recall indicating a very poor separation of ID an OOD pixels.
%The softmax histogram uses logarithmic scale. 

\begin{figure}[htb]
  \centering
  \includegraphics[width=1\columnwidth]{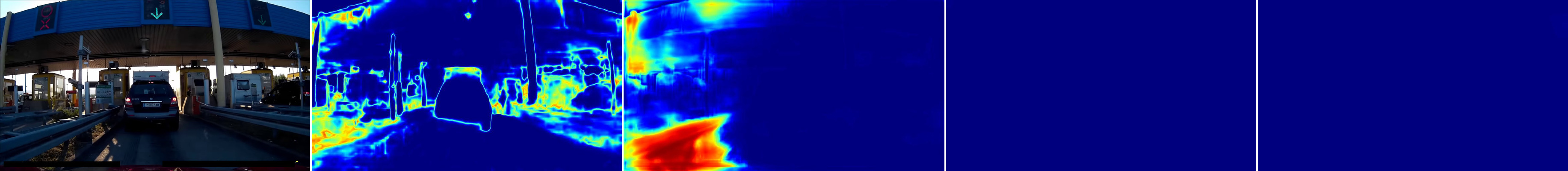}
  \includegraphics[width=1\columnwidth]{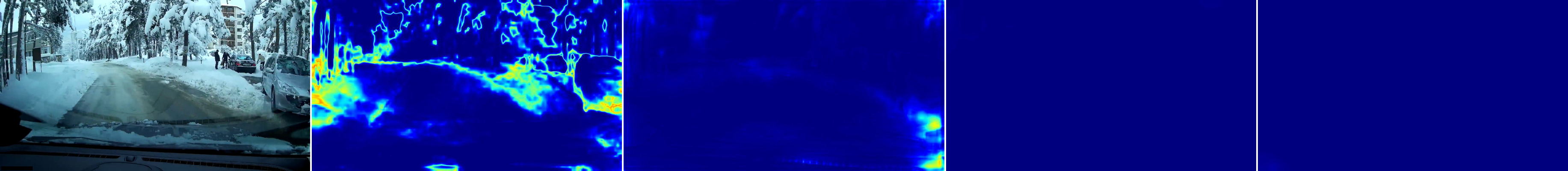}
  \includegraphics[width=1\columnwidth]{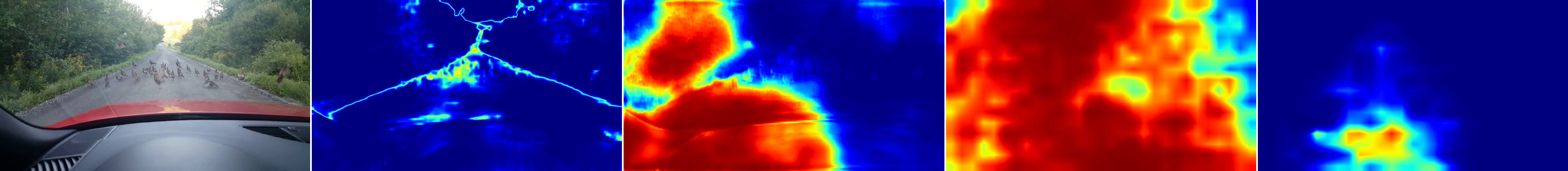}
  \includegraphics[width=1\columnwidth]{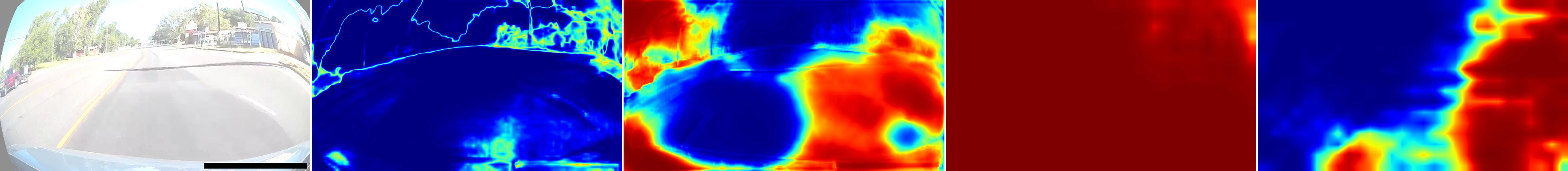}
  \caption{OOD detection 
    in in-distribution WildDash test images. 
    We use the same column arrangement
    as in Figure\ \ref{fig:ID-as-OOD}.
    The row three is interesting because it contains animals
    which are classified as ID by the model trained on WildDash val.
    This is likely due to the fact that WildDash val 
    contains images with animals on the road 
    (although not ducks).
    Notice finally that the model instance trained on WildDash val
    performs better on distorted images.  
  }
  \label{fig:ID}
\end{figure}

\begin{figure}[htb]
  \centering
  \includegraphics[width=1\columnwidth]{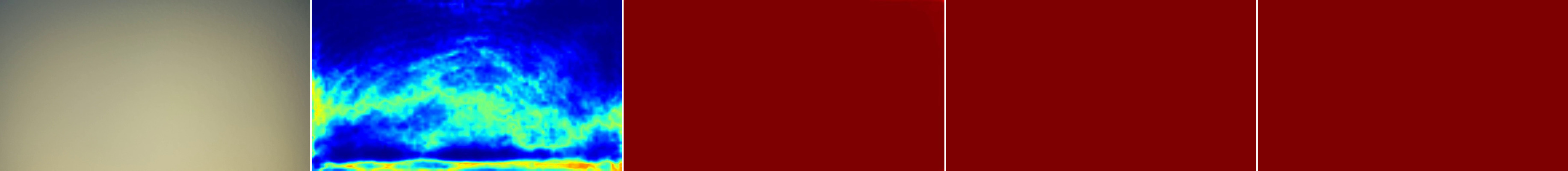}
  \includegraphics[width=1\columnwidth]{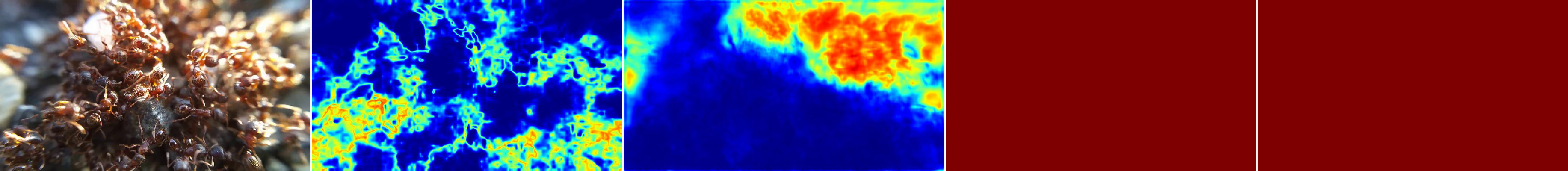}
  \includegraphics[width=1\columnwidth]{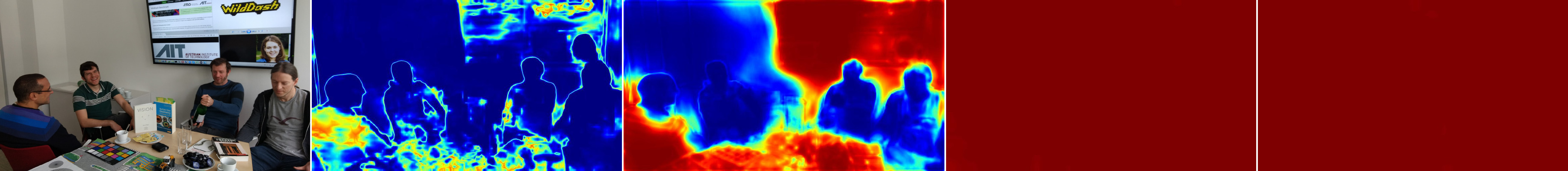}

  \caption{OOD detection 
    in negative WildDash test images. 
    We use the same column arrangement
    as in Figure\ \ref{fig:ID-as-OOD}.
    The ROB model classifies 
    all non-indoor images as ID, 
    different from the models 
    that have seen the ILSVRC dataset.
  }
  \label{fig:OOD}
\end{figure}

We show the confidence that the corresponding pixel is OOD 
in the second column of Figures \ref{fig:ID-as-OOD}, 
\ref{fig:ID} and \ref{fig:OOD},
where red denotes higher confidence.
We see that OOD responses obtained by the max-softmax approach
do not correlate with epistemic uncertainty.
Instead they occur on semantic borders, small objects 
and distant parts of the scene, that is on details 
which occur in most training images.

Table \ref{table:APconf} shows the OOD detection 
with the augmented primary model and trained confidence. 
This approach achieves a better mIoU performance than 
the basic segmentation model trained only on Cityscapes. 
This suggests that training with uncertainty alleviates overfitting. 
Still, the uncertainty estimation itself 
proves to be the worst predictor of OOD pixels 
among all approaches considered in our experiments. 
This suggests that the confidence head is "confident" by default 
and must be taught to recognize the pixels which should be uncertain. 
Since this model performed poorly we do not show 
its precision-recall curve nor its OOD segmentation results.

\setlength{\tabcolsep}{4pt}
\begin{table}[htb]
\begin{center}
\caption{Average precision for OOD detection 
    by the augmented primary model 
    featuring a trained confidence head.}
\label{table:APconf}
\begin{tabular}{|c||cc|cc|}
  \hline
  Training set
  & \multicolumn{2}{c|}{WD test selection}  
  & \multicolumn{2}{c|}{WD test complete} 
\\
  \cline{2-5}
  & max-softmax
  & confidence
  & max-softmax
  & confidence
\\
 \hline
 \hline
  \multicolumn{1}{|l||}{City} & 10.61 & 9.52 & 15.62 & 11.38\\
\hline
\end{tabular}
\end{center}
\end{table}
\setlength{\tabcolsep}{1.4pt}

Table \ref{table:APROB} shows OOD detection results 
for a model that has seen both indoor scenes 
and road-driving scenes by training on 
all four datasets from the ROB challenge. 
There is a significant jump in average precision 
between this model and approaches trained 
only on road-driving scenes. 
Interestingly, this model recognizes most pixels 
in the five artificially altered negative WildDash test images as ID 
(column 3, Figure ~\ref{fig:ID-as-OOD}). 
The model works well on ID images (Figure \ref{fig:ID}),
however it makes some errors in some OOD images. 
The third column of Figure \ref{fig:OOD} shows 
that some pixels like ants (row 2) 
are recognized as ID samples. 
Interestingly, the model recognizes pixels at people as ID,
even though they are located in an OOD context 
(row 3 in image \ref{fig:OOD}).

We also show the precision-recall curve for this model in
the second column of Figure~\ref{fig:hists}. The precision is
high for low values of recall. Furthermore, the precision
remains high along a greater range of recall values when
using probabilities for OOD detection.

\setlength{\tabcolsep}{4pt}
\begin{table}[htb]
\begin{center}
\caption{Average precision for OOD detection with the classifier
  trained on all four datasets from the ROB 2018 challenge:
  Cityscapes trainval, KITTI train, WildDash val and ScanNet train.
  Note that OOD predictions correspond to indoor classes 
  from the ScanNet dataset.}
\label{table:APROB}
\begin{tabular}{|c||c|c|}
  \hline
  Training set
  & \multicolumn{1}{c|}{WD test selection}  
  & \multicolumn{1}{c|}{WD test complete} 
 \\
 \hline
 \hline
  \multicolumn{1}{|l||}{ROB 2018} & 69.19 & 54.71\\
\hline
\end{tabular}
\end{center}
\end{table}
\setlength{\tabcolsep}{1.4pt}

Table \ref{table:APbin} shows average precision 
for the proposed discriminative OOD detectors 
which jointly train on the ILSVRC dataset and
road-driving imagess from different datasets. 
We start with the model instance which was trained
using only Cityscapes images as ID examples. 
Interestingly, this instance performs poorly 
because it classifies all WildDash test images as OOD. 
This result indicates that Cityscapes dataset 
encourages overfitting due to all images 
being acquired with the same camera. 
The other two instances of the model perform better 
than all other approaches presented in this paper. 
The model instance trained on Vistas 
performs significantly better 
than the model instance trained only on Cityscapes. 
Still, we obtain the best results with the instance 
that has seen WildDash validation set.
This suggests that a model that needs to work in challenging environments 
also needs to see challenging examples during training.
Interestingly, these two instances recognize 
pixels from artificially altered images as ID samples,
which is evident from the drop of performance 
between the two columns in Table \ref{table:APbin}
as well as from columns 4 and 5 in Figure \ref{fig:ID-as-OOD}. 
Finally, these models do not recognize ID classes in OOD context: 
people sitting in a room are classified as OOD samples
as shown in row 3 in Figure \ref{fig:OOD}.

\setlength{\tabcolsep}{4pt}
\begin{table}[htb]
\begin{center}
\caption{Average precision for discriminative OOD detection
  on the WildDash test dataset.
  The OOD detection model is jointly trained on ILSVRC (OOD pixels)
  and road-driving images from different datasets (ID pixels).
}
\label{table:APbin}

\begin{tabular}{|c||c|c|}
  \hline
  Training set
  & \multicolumn{1}{c|}{WD test selection}  
  & \multicolumn{1}{c|}{WD test complete} 
 \\
 \hline
 \hline
  \multicolumn{1}{|l||}{city,img}    & 32.11 & 24.83 \\ 
  \hline
  \multicolumn{1}{|l||}{city,wd,img} & 96.24 & 71.43 \\
  \hline
  \multicolumn{1}{|l||}{vistas,img}  & 89.23 & 67.44 \\
  %\hline
  %bin & vistas,wd,img & 86.71& 70.05 & 87.27 & 70.56 \\
  %bin-attention & vistas,wd,img & 66.44 & 5.36 & 57.53 & 5.36 & 5.36 & N/A \\
  \hline
\end{tabular}
\end{center}
\end{table}
\setlength{\tabcolsep}{1.4pt}

Precision-recall curves for the model instance trained on Vistas 
and the model instance trained on Cityscapes + WildDash 
can be seen in Figure\ \ref{fig:hists}, in columns 3 and 4 respectively. 
The curve for the model that has only seen the Vistas dataset slopes 
relatively constantly, while the curve for the model that 
has seen the WildDash validation set remains constant and high and then drops suddenly. 
This drop is due altered valid images shown in Figure \ref{fig:ID-as-OOD}.

\begin{figure}[htb]
  \centering
  \includegraphics[width=0.9\columnwidth]{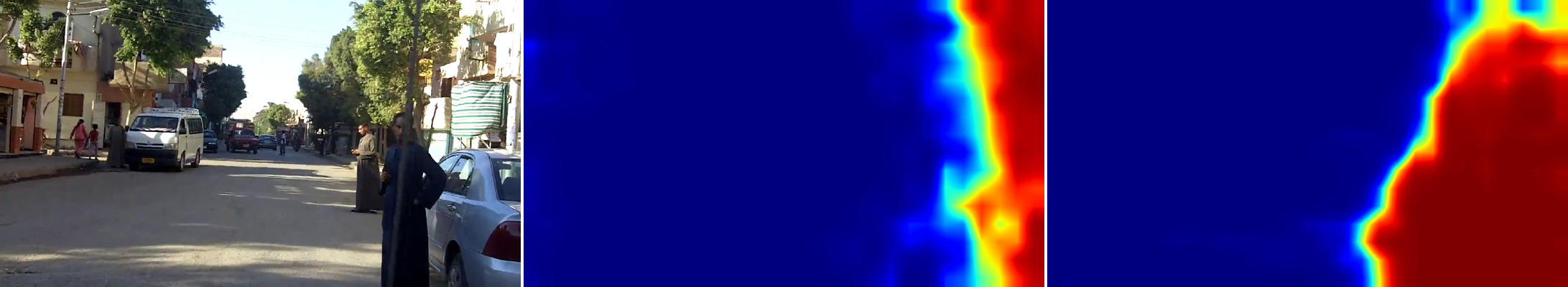}
  \includegraphics[width=0.9\columnwidth]{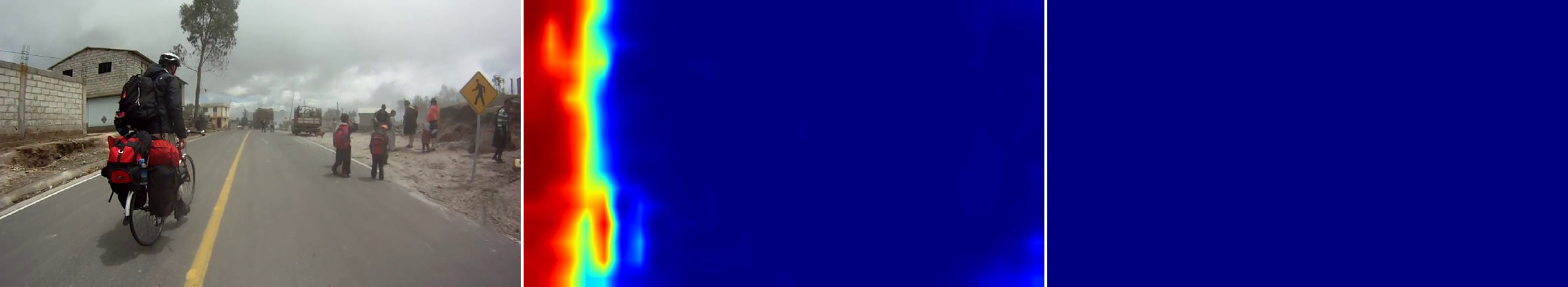}
  \includegraphics[width=0.9\columnwidth]{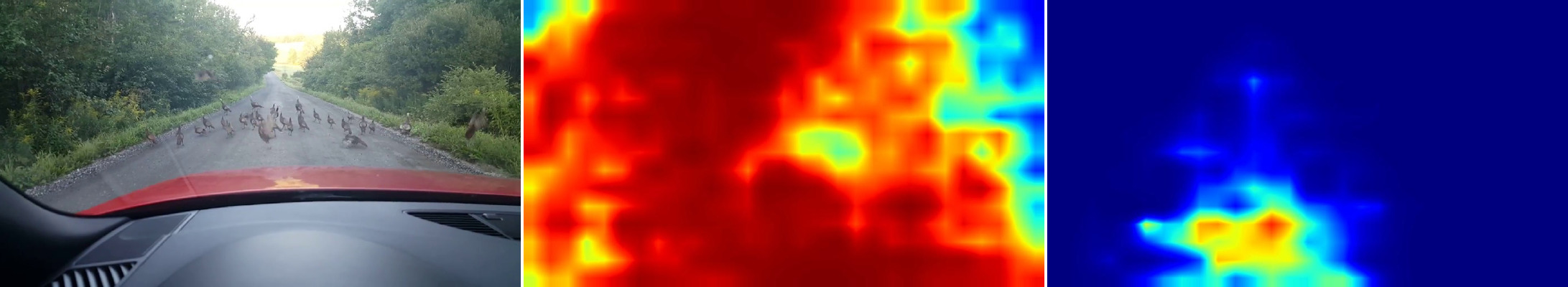}
  \includegraphics[width=0.9\columnwidth]{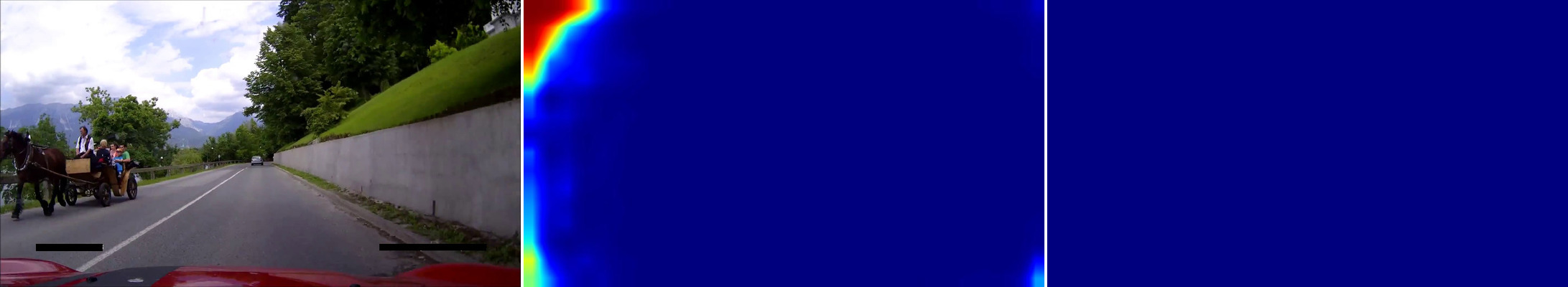}
  \caption{Examples of OOD pixel detections
    in positive WildDash test images. 
    The columns correspond to:
    i) original image, 
    ii) discriminative OOD detection trained on Vistas, 
    and iii) discriminative OOD detection 
    trained on Cityscapes and WildDash val.
    Red denotes the confidence 
    that the corresponding pixel is OOD
    which can be interpreted as epistemic uncertainty.
  }
  \label{fig:mistakes}
\end{figure}

Finally, we show a few difficult cases in Figure\ \ref{fig:mistakes} 
to discuss the space for improvement. 
Rather than visualizing the classification (which has sharp edges), 
we show confidence that the pixel is OOD. 
The first two rows contain images which are clearly inliers,
however our discriminative models suspect 
that some of their parts are OOD. 
This is probably caused by existence of ID classes 
in ILSVRC images (e.g. caravan, sports car, dome). 
Our models are not expressive enough to indicate 
which ILSVRC classes caused these errors.
Images in rows 3 and 4 are ID images that contain OOD objects, in this case animals. 
Future models would benefit by better considering
the overlap between ILSVRC and the primary training dataset.
Furthermore, the predictions are very coarse
due to image-wide annotations of the training datasets.
Finer pixel-level predictions would likely be obtained
by training on images that contain both ID and OOD pixels.

\subsection{Results on other datsets}

Table \ref{table:APbinOther} shows
that the model instance trained on 
Vistas and ILSVRC generalizes
very well for OOD detection in 
PASCAL and Cityscapes images.
On the other hand, 
the model trained on Cityscapes 
recognizes most Vistas pixels as OOD.
This shows that the Cityscapes dataset
is inappropriate for training
generic models for understanding
road-driving scenes.

\setlength{\tabcolsep}{4pt}
\begin{table}[htb]
\begin{center}
\caption{Pixel accuracy of discriminative OOD detection on various datasets.
PASCAL$^*$ denotes PASCAL VOC 2007 trainval
without Cityscapes classes
(bicycle, bus, car, motorbike, person, train).}
\label{table:APbinOther}
\begin{tabular}{|l||l|c|}
\hline
  Training set & 
  Test set &
  OOD incidence
 \\
  \hline
  \hline
  Vistas, ILSVRC &
  Cityscapes test &
  0.01\%
 \\
 \hline
  Vistas, ILSVRC &
  PASCAL$^*$ &
  99.99\%
 \\
 \hline
  Cityscapes, ILSVRC &
  Vistas val &
  93.76\%
  \\
  \hline
\end{tabular}
\end{center}
\end{table}
\setlength{\tabcolsep}{1.4pt}

\section{Conclusion}

% We need better datasets
Graceful performance degradation in presence of 
unforeseen scenery is a crucial capability 
for any real-life application of computer vision.
Any system for recognizing images in the wild 
should at least be able to detect such situations
in order to avoid disasters and fear of technology.

This paper presented a novel approach for detecting 
out-of-distribution (OOD) pixels in test images.
The main idea is to recognize the outliers 
as being more similar to some general dataset such as ILSVRC 
than to the training dataset 
of the primary model for the specific vision task. 
In order to implement that idea, we train the OOD detection model 
on much more training data than the primary model.

Experiments show substantial improvement 
of OOD detection AP performance 
with respect to all previous approaches. 
The proposed method is able to distinguish 
negative WildDash images from the positive ones.
%It is also able to deliver fair recognition performance 
%in detecting animals in traffic scenes
%with a model that was not specifically trained for animal detection.
The presented results emphasize the need 
for more comprehensive background datasets,
as well as for including well-annotated 
out-of-distribution samples to task-specific datasets.

Future work will address employing these results 
as a guide for better direction of the annotation effort
as well as towards further development of approaches
for recognizing epistemic uncertainty in images and video.

\clearpage
\section*{Acknowledgement}
This work has been partially supported by Croatian Science foundation. We would like to thank Josip Krapac and Axel Pinz for useful comments on this draft. We would like to thank Ivan Grubi{\v s}i{\' c} for performing baseline experiments with epistemic uncertainty based on Monte Carlo dropout.

\bibliography{iclr2019_conference}
\bibliographystyle{iclr2019_conference}
\clearpage

\begin{appendices}

\section{Dense OOD detection in images with mixed content}

The Wilddash dataset is the only 
publicly available dataset 
that provides OOD images. 
Unfortunately, the Wilddash OOD content
is annotated only on the image level.
This makes Wilddash unsuitable for testing 
detection of unfamiliar objects in familiar settings.
We therefore propose six new datasets for that purpose.

We also propose an improved training procedure
which allows the proposed discriminative OOD detection model 
to accurately predict borders of OOD objects.
This procedure is used to train a new instance 
of the discriminative OOD model 
which is going to be evaluated in experiments.

Finally, we present and discuss experiments which compare 
the AP performance across several OOD models and datasets. 

\begin{comment}
There are no images with annotated 
OOD pixels in ID scenarios, 
meaning there are no images that contain
traffic scenes with unexpected objects.
There are a few images that contain traffic scenes with animals, 
but those animals are ignored during evaluation.
One of the motivations for dense OOD prediction is to be able to
detect unfamiliar objects in familiar settings.
\end{comment}

\subsection{Test datasets}
\label{Sets}

In order to be able to evaluate 
how different OOD detection methods 
perform when OOD pixels are present in ID scenes,
we create six new datasets.
Three of these datasets include images
which contain both ID and OOD pixels.
We shall use these datasets for evaluating
various OOD detection approaches.
The remaining three datasets 
are designed for control experiments
in which we shall explore whether 
the evaluated OOD detection approaches
are able to react to pasted ID content.

We obtain the first two datasets 
by pasting Pascal VOC 2007 animals 
of different sizes to images from Vistas val.
Pascal was chosen because it contains 
many densely annotated object instances
which are out-of-distribution for road-driving scenes.
We do not use Vistas train at this point
since we wish to preserve it for OOD training.
The three control datasets are formed
by pasting objects across images
from road driving datasets.
The final dataset contains a selection of Vistas images
in which a significant number of pixels are labeled 
as the class 'ground animal'. 
The last dataset is the closest match to a real-world scenario 
of encountering an unexpected object while driving,
however, the number of its images is rather small
(hence the need for datasets obtained by pasting).

\begin{comment}
To evaluate how different OOD detection methods perform 
when OOD pixels are present in ID scenes, 
we create a new dataset by pasting OOD objects
into images from road-driving datasets.
We start from around 6000 images from Pascal VOC 2007
with pixel level annotations 
for the semantic segmentation task.
We pick all instances of 
consider image regions corresponding to 
Pascal VOC 2007 animals and paste them 
into Vistas images. 
To ensure that the model does not detect OOD patches
because of different conditions under which Pascal images were
take (type of camera, resolution, lighting conditions), additional
sets were created which pasted ID objects into ID images.

The final set was created by using Vistas images which contain
the class 'ground animal'. 
This set is the closest match to a real
world scenario of encountering unexpected objects.

The following sections describe these sets in more detail.
\end{comment}

\subsubsection{Pascal to Vistas 10\%}

\begin{figure}[htb]
  \centering
  \includegraphics[width=1\columnwidth]{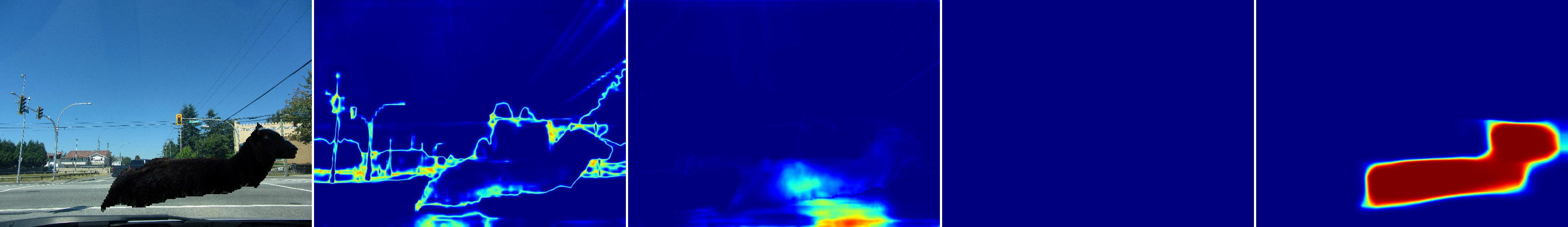}
  \includegraphics[width=1\columnwidth]{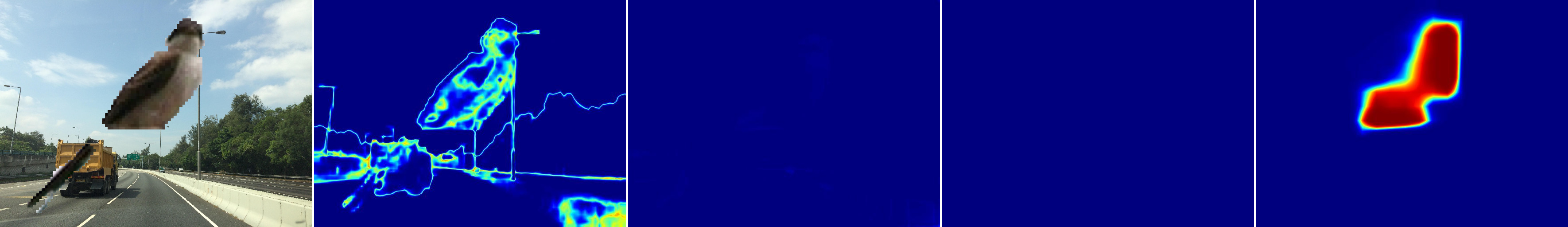}
  \includegraphics[width=1\columnwidth]{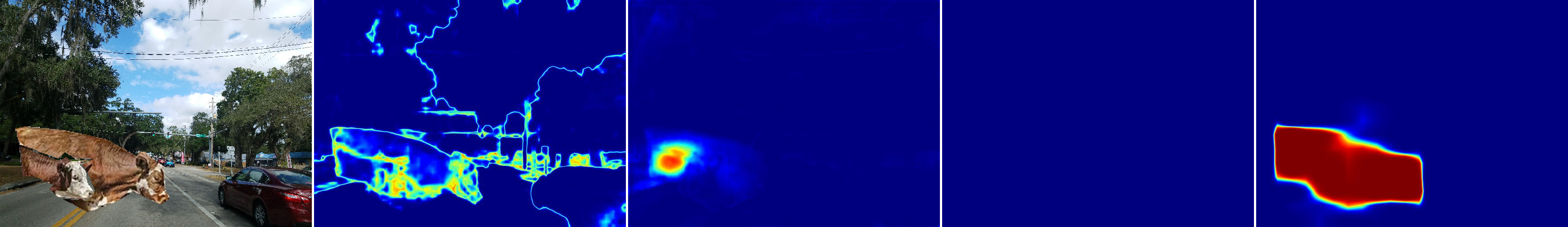}

  \caption{OOD detection in Vistas images with pasted
  	Pascal objects that take up at least 10\% of the image.
    The columns correspond to:
    i) original image, 
    ii) max-softmax of the primary model
        (cf.\ Table \ref{table:APseg}), 
    iii) OOD detection with the ROB model
        (cf.\ Table \ref{table:APROB}), 
    iv) discriminative OOD detection trained 
        on entire images from ILSVRC (OOD) 
        and Vistas train (ID)
        (cf.\ Table \ref{table:APbin}), 
    and v) 
      discriminative OOD detection 
      trained on entire ILSVRC images (OOD),
      and ILSVRC bounding boxes (OOD)
      pasted over Vistas images without ground animals (ID).
    Red denotes the confidence 
    that the corresponding pixel is OOD,
    which can be interpreted as epistemic uncertainty.
    Max-softmax of the primary model detects borders.
    The model trained according to \ref{ss:traindata}
    manages to accurately detect the OOD shape. 
    The ROB model manages to detect the
    position the pasted patch, while the 
  	discriminative model trained only on the whole 
    OOD images does not detect 
    any of the pasted patches.
  }
  \label{fig:pascal_vistas_10}
\end{figure}

We start by locating Pascal images 
with segmentation groundtruth
which contain any of the 7 animal classes: 
bird, cat, cow, dog, horse and sheep.
We select 369 large Pascal objects 
from their original images
using pixel-level segmentation groundtruth.
For each selected object we choose 
a random image from Vistas val,
resize the object to cover at least 10\% image pixels
and then paste the object at random image location.
This results in 369 combined images.
Examples of the combined images are shown
in column 1 in Figure \ref{fig:pascal_vistas_10}.

\begin{comment}
We started the evaluation by pasting objects form Pascal VOC 2007
into Vistas images. 
Pascal was chosen because it contains densely
annotated object instances. 
Furthermore, it is out-of-distribution
for traffic scenes, but also different from Imagenet, and is a good
indicator if a model can distinguish between ID and OOD pixels, 
even if those OOD pixels are not from the OOD distribution it
was trained on.

The objects chosen to be pasted belong to the Pascal VOC
animal classes: bird, cat, cow, dog, horse and sheep.
These objects were cut from the original image, resized to 
take up at least 10\% of the target Vistas validation image using
bilinear interpolation, and pasted to a random positions into a 
random Vistas image.

This resulted in 369 images. Column 1 in Figure \ref{fig:pascal_vistas_10} shows
examples of resulting images.
\end{comment}

\subsubsection{Pascal to Vistas 1\%}

\begin{figure}[htb]
  \centering
  \includegraphics[width=1\columnwidth]{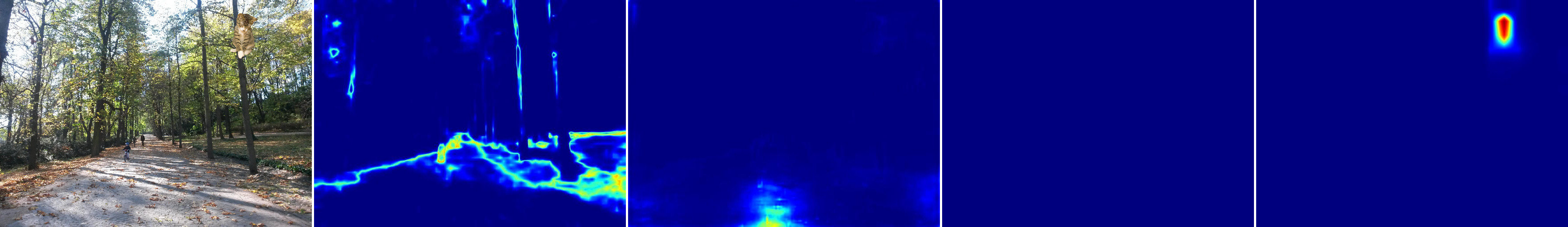}
  \includegraphics[width=1\columnwidth]{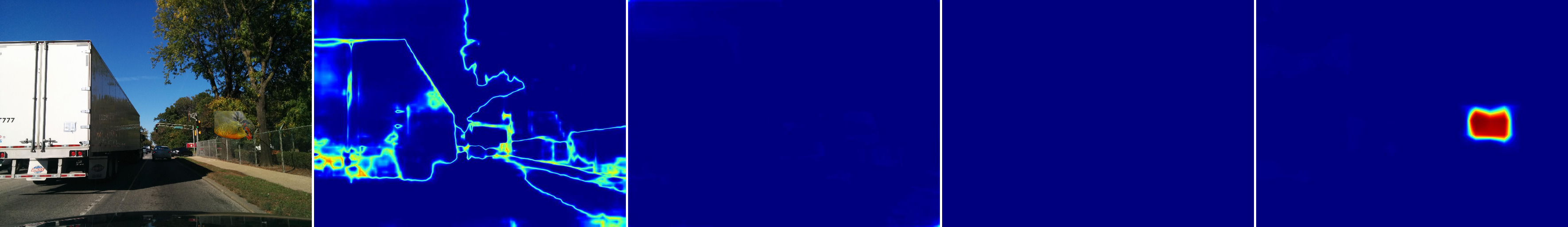}
  \includegraphics[width=1\columnwidth]{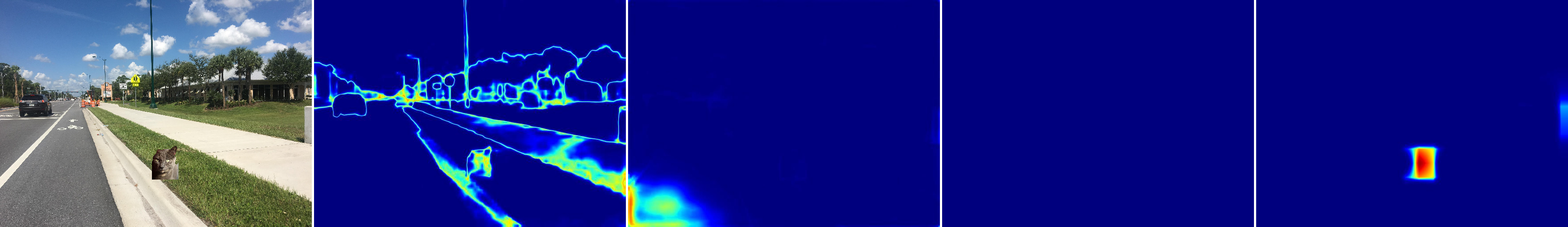}

  \caption{OOD detection in Vistas images with pasted
  	Pascal objects that take at least 1\% of the image.
    We use the same column arrangement and colors
    as in Figure\ \ref{fig:pascal_vistas_10}.
    The model trained as described in \ref{ss:traindata}
    is able to detect even the relatively
    small objects pasted into a similar background.
    The ROB model fails to detect 
    the location of the pasted patch.
  }
  \label{fig:pascal_vistas_1}
\end{figure}

A possible issue with resizing objects before pasting 
might be that the OOD model may succeed 
to detect the pasted objects 
by recognizing the resizing artifacts 
instead of the novelty.
In order to address this issue, we form another dataset as follows.
We iterate over all instances of Pascal objects,
we choose a random image from Vistas val
and paste the object without any resizing 
only if it takes at least 1\% image pixels.
This results in 31 combined images.
This datasets is more difficult than the previous one
since OOD patches are much smaller. 
Examples can be seen in the first column 
of Figure \ref{fig:pascal_vistas_1}.

\begin{comment}
A possible issue with resizing objects before pasting them
might be that the OOD model detects pasted objects as OOD
because it actually detects preprocessing artifacts instead of
OOD pixels.
  
To test if the model detect OOD patches that have not been
preprocessed, Pascal object instances that take up at least 1\% of
the randomly chosen target Vistas validation image were cut from the
original images and pasted into a target image without preprocessing.

This results in 31 images that have smaller OOD patches than the                                                     
previous set. Examples can be seen in the first column Figure
\ref{fig:pascal_vistas_1}
 
\end{comment}

\subsubsection{City to City}

\begin{figure}[htb]
  \centering
  \includegraphics[width=1\columnwidth]{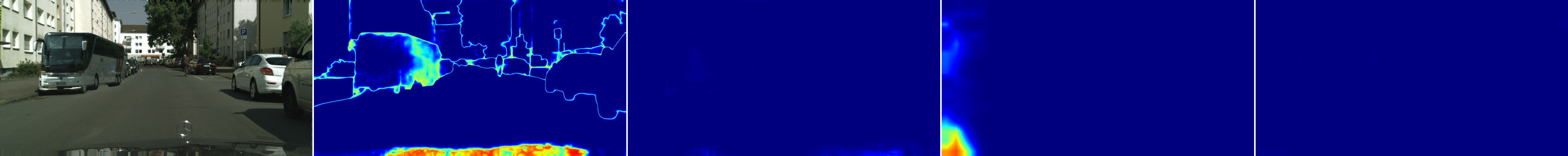}
  \includegraphics[width=1\columnwidth]{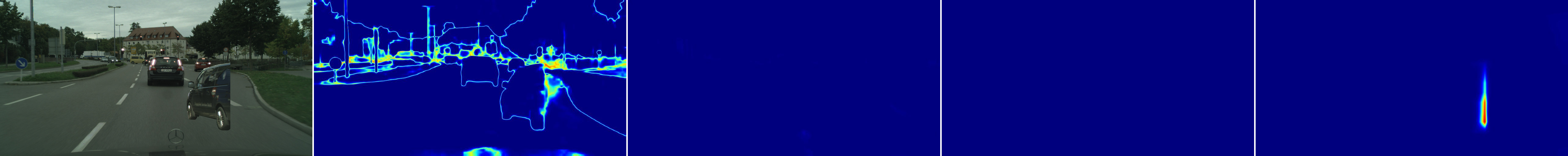}
  \includegraphics[width=1\columnwidth]{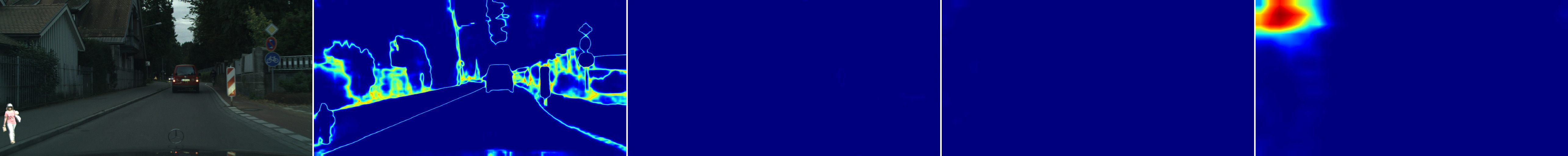}

  \caption{OOD detection in Cityscapes images with pasted
  	Cityscapes instances that take at least 0.5\% of the image.
    We use the same column arrangement and colors
    as in Figure\ \ref{fig:pascal_vistas_10}.
    None of the models accurately detect the pasted patches. 
    The fourth model seems to react to 
    borders of pasted content (row 2).
  }
  \label{fig:city_to_city}
\end{figure}

We create this dataset by pasting 
a random object instance Cityscapes val
at a random location of a different 
random Cityscapes validation image.
The only condition is that the object instance 
takes at least 0.5\% of the cityscapes image. 
No preprocessing is performed before the pasting.
Performance on this set indicates 
whether a model detects OOD pixels 
due to different imaging conditions 
in which the patches were acquired.
This dataset contains 288 images.
Examples can be seen in the first
column Figure \ref{fig:city_to_city}.

\begin{comment}
City to city set was created by pasting randomly selected object
instance from a Cityscapes validation image to another randomly
selected Cityscapes validation image at a random location within
the images, with the only condition being 
that the object instance has to take up at least 0.5\% of the 
cityscapes image. No preprocessing was done before pasting.

Performance on this set should indicate if the model detects
OOD pixels because of different shooting conditions in which the 
patches were shot.

This set contains 288 images. Examples can be seen in the first
column Figure \ref{fig:city_to_city}.
\end{comment}

\subsubsection{Vistas to City}

\begin{figure}[htb]
  \centering
  \includegraphics[width=1\columnwidth]{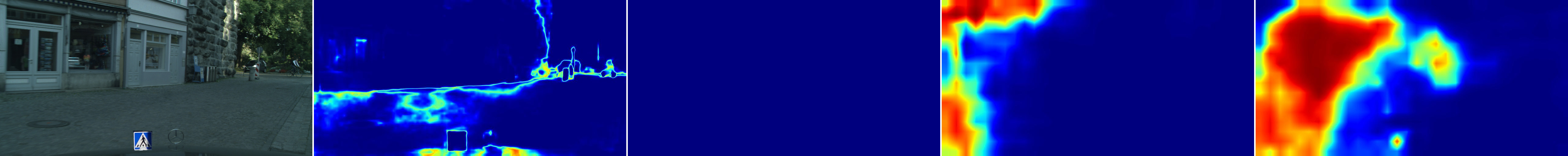}
  \includegraphics[width=1\columnwidth]{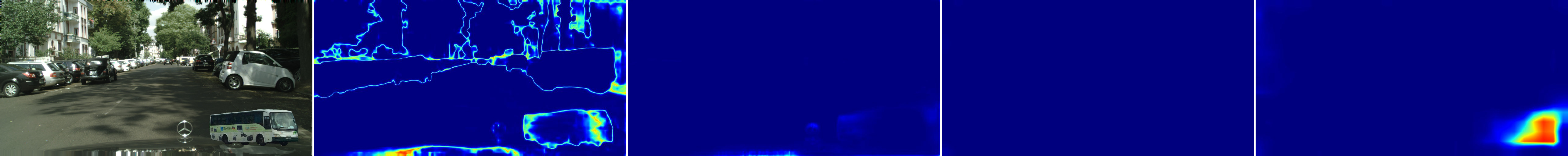}
  \includegraphics[width=1\columnwidth]{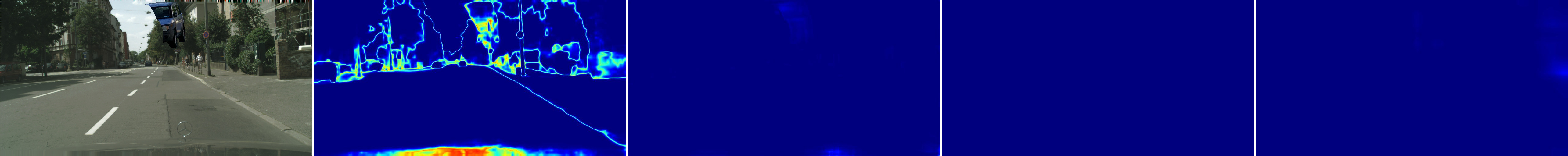}

  \caption{OOD detection in Cityscapes images with pasted
  	Vistas instances that take at least 0.5\% of the image.
    We use the same column arrangement and colors
    as in Figure\ \ref{fig:pascal_vistas_10}.
    The fourth model experiences trouble with 
    atypical Cityscapes images (row 1)
    and detects borders of the pasted patches.
  }
  \label{fig:vistas_to_city}
\end{figure}

We create this dataset by pasting 
a random object instance from Vistas val 
into a random image from Cityscapes val. 
The pasted instance has to take 
at least 0.5\% of the Cityscapes image. 
No preprocessing is performed before the pasting.
Performance on this set indicates 
whether the model is able to detect
different camera characteristics of the patch 
rather than real OOD pixels.
This dataset contains 1543 images.
Some examples are shown in 
Figure \ref{fig:vistas_to_city}.

\begin{comment}
Vistas to city set was created by pasting a randomly selected object
object instance from a Vistas validation image and randomly pasting it into
a randomly selected Cityscapes validation image. The Vistas object
instance had to take up at least 0.5\% of the Cityscapes image. No 
preprocessing was done before pasting.

Performance on this set should indicate if the model detects
different camera characteristics of the patch rather than real OOD
pixels.

This set contains 1543 images. The first column in Figure 
\ref{fig:vistas_to_city} shows some examples of vistas to city 
images.
\end{comment}

\subsubsection{Self to self}

\begin{figure}[htb]
  \centering
  \includegraphics[width=1\columnwidth]{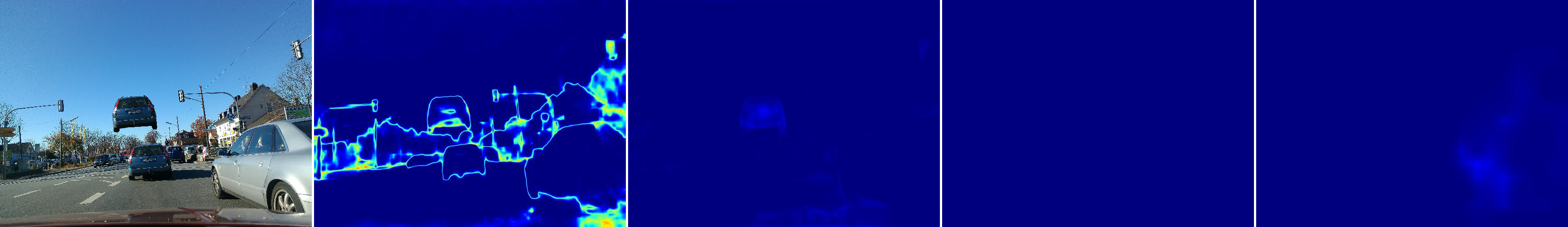}
  \includegraphics[width=1\columnwidth]{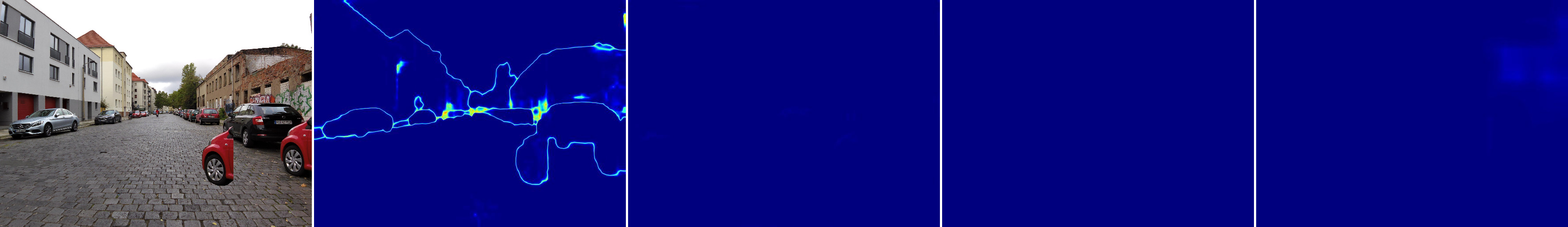}
  \includegraphics[width=1\columnwidth]{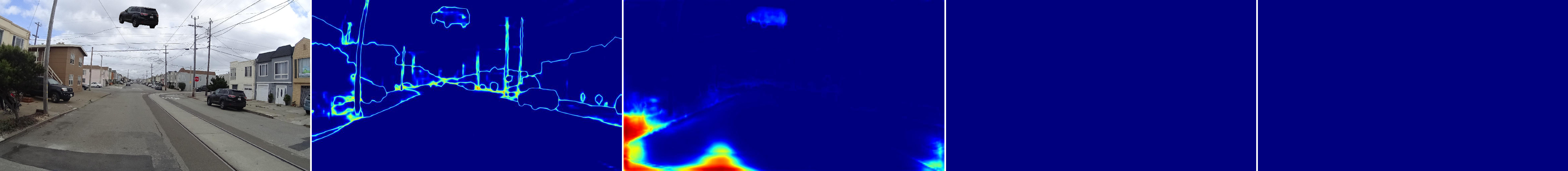}

  \caption{OOD detection in Vistas images that contain
   	objects copied and pasted from the same image
    We use the same column arrangement and colors
    as in Figure\ \ref{fig:pascal_vistas_10}.
    None of the models detect the pasted patches.
  }
  \label{fig:self_to_self}
\end{figure}

We create this dataset by pasting 
a randomly selected object instance from a Vistas image 
to a random location in the same image. 
The object instance had to take at least 0.5\% 
of the vistas image.
No preprocessing was performed before the pasting.
Performance on this set indicates whether 
the model is able to detect objects at
unusual locations in the scene.
This set contains 1873 images.
Some examples can be seen in 
Figure \ref{fig:self_to_self}.

\begin{comment}
this set was created by pasting a randomly selected object instance 
from a Vistas image to a random location in the same image. The
object instance had to take up at least 0.5\% of the vistas image.
No preprocessing was done before pasting.

Performance on this set indicates if the model detects objects at
unusual locations in the scene.

This set contains 1873 images which can be seen in Figure
\ref{fig:self_to_self}.
\end{comment}

\subsubsection{Vistas animals}

\begin{figure}[htb]
  \centering
  \includegraphics[width=1\columnwidth]{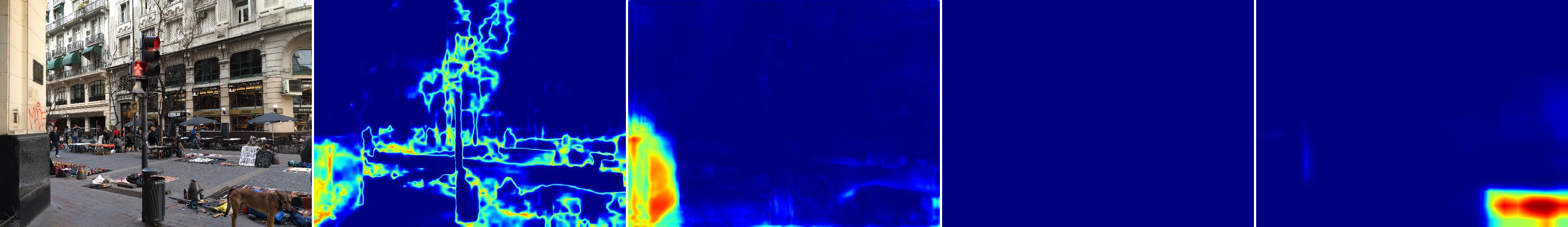}
  \includegraphics[width=1\columnwidth]{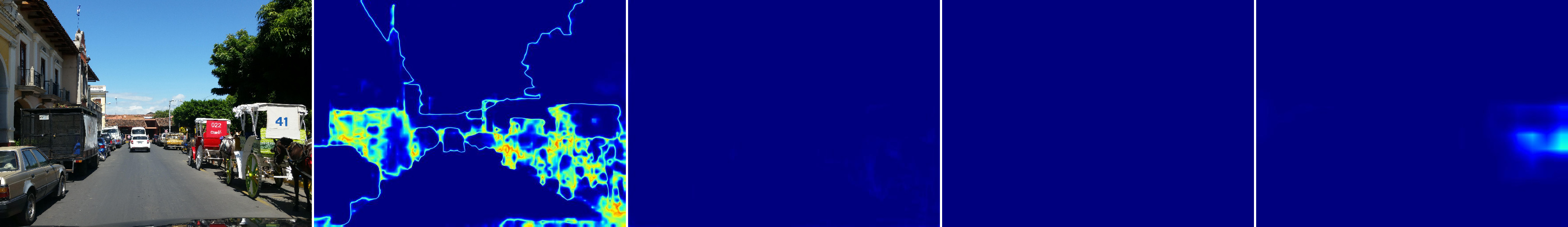}
  \includegraphics[width=1\columnwidth]{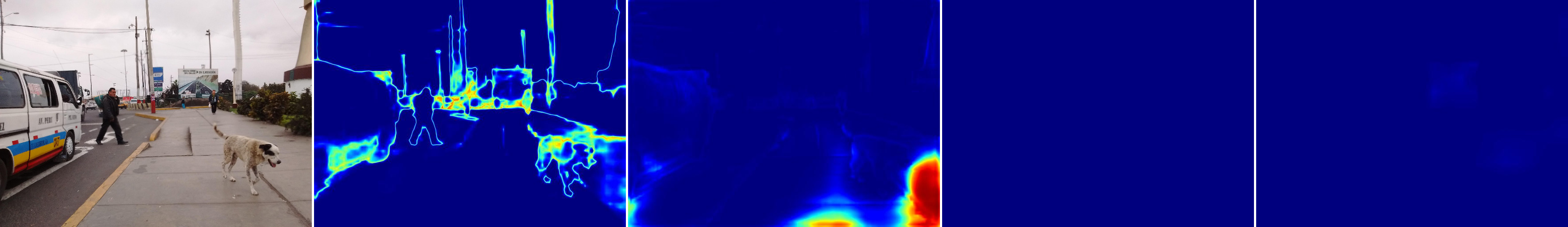}
  \caption{OOD detection in Vistas images 
    that contain the class 'ground animal'.
    We use the same column arrangement and colors
    as in Figure\ \ref{fig:pascal_vistas_10}.
    Only the fourth model manages 
    to accurately segment an animal in row 1, 
    and reacts to animals in other two images. 
    The ROB model detects some parts as OOD 
    however those regions do not 
    correspond to animal locations.
  }
  \label{fig:animals_vistas}
\end{figure}

This dataset is a subset of Vistas 
training and validation images
which contain instances labeled 'ground animal' 
that take at least 0.7\% of the image.
This set is closest to the real-world scenario 
of encountering unknown objects 
in ID road driving scenes. 
Unlike in images with pasted Pascal animals, 
OOD detection is unable to succeed 
by recognizing the pasting artifacts 
or different imaging conditions.
This set contains 8 images. 
Three of those are shown in the first column 
of Figure \ref{fig:animals_vistas}.

\begin{comment}
Vistas animals is a subset of Vistas training and validation images
which contain patches of pixels labeled 'ground animal' that take up
at least 0.7\% of the image.

This set is the closest thing to a real world scenario of
encountering unknown objects in ID scenes. Most of the segmentation
models are currently not trained to recognize all of the cityscapes
classes (meaning animals are usually ignored during training and 
evaluation), so it can assumed that animals can currently be
considered OOD in the context of traffic scene semantic segmentation.

Furthermore, unlike images with pasted Pascal patches, there is no
chance that OOD pixels are detected because the OOD patches come
from images taken in different shooting conditions.

This set contains 8 images. Three of those can bee seen in the first column of Figure \ref{fig:animals_vistas}.
\end{comment}

\subsection{Training dataset}
\label{ss:traindata}

In order to be able to cope with images 
containing both ID and OOD pixels,
we perform the following changes to the training dataset.
First, we remove all Vistas images which contain
instances of the class 'ground animal'
from the training split, 
regardless of the instance size.
Then, we select 544\,546 ILSVRC images 
in which bounding box annotation is available.
Each of the selected ILSVRC images 
is used only once during training,
either as i) a single image or ii) a combined image 
obtained by pasting the resized bounding box 
to a random location of a random Vistas train image.
In the former case, the bounding box is labeled as OOD 
while the rest of ILSVRC image is ignored.
In the latter case, the bounding box 
is resized to contain 5\% pixels of the Vistas image,
the resulting ILSVRC pixels are labeled as OOD 
and the rest is labeled as ID.
In both cases the resulting image is resized 
so that the shorter dimension equals 512 pixels
and randomly cropped to the resolution of 512$\times$512. 

\subsection{Model and training}

We use the same fully convolutional discriminative OOD model 
as described in Section \ref{sec:models}.
The model is trained according to procedure 
described in Section \ref{sec:train},
except that the training dataset 
is composed as described in \ref{ss:traindata}.

\pagebreak
\subsection{Experimental results}

Tables \ref{table:APanimals} and \ref{table:APcontrol} show
the average precision performance
for all OOD models described in the paper,
together with the new instance 
of the discriminative model which is trained 
on the train dataset described in 
Section \ref{ss:traindata} of this appendix.
The models are evaluated on all six test datasets 
presented in section \ref{Sets}.
We compare the achieved performance 
with the respective results on the WildDash test dataset
which we copy from Section \ref{sec:exp}.

\setlength{\tabcolsep}{4pt}
\begin{table}[htb]
\begin{center}
\caption{Average precision for discriminative OOD detection 
  on the test datasets with images 
  that have both ID and OOD pixels. 
  Labels stand for max-softmax of the primary model (ms),
  max-softmax of the model with trained confidence (ms-conf),
  primary model trained for the ROB challenge (ROB),
  and the discriminative model (discrim).
}
\label{table:APanimals}
\begin{tabular}{|l|l||c|c|c|c|}
\hline
  Model &
  Training set & 
  PascalVistas10 &
  PascalVistas1 &
  VistasAnimals &
  Wilddash selection 
 \\
  \hline
  \hline
  ms &
  city &
  28.81 &
  9.91 &
  6.05 &
  10.09
  
 \\
 \hline
  ms &
  city, wd &
  34.27 &
  8.8 &

  6.79 &
  17.62
 \\
 \hline
  ms-conf &
  city &
  26.09 &
  8.04 &

  5.94 &
  10.61
  \\
  \hline
  ROB &
  ROB &
  25.65 &
  4.55 &

  2.96 &
  69.19
  \\
  \hline
  discrim &
  city, img &
  34.07 &
  3.19 &

  2.28 &
  32.11
  \\
  \hline
  discrim &
  city, wd, img &
  24.46 &
  3.19 &

  4.59 &
  \textbf{96.24}
  \\
  \hline
  discrim &
  vistas, img &
  13.14 &
  2.39 &

  2.4 &
  89.23
  \\
  \hline
  discrim &
  vistas-a, img\_bb &
  \textbf{87.87} &
  \textbf{78.58} &

  \textbf{25.61} &
  68.59
  \\
  \hline
\end{tabular}
\end{center}
\end{table}
\setlength{\tabcolsep}{1.4pt}

Images \ref{fig:pascal_vistas_10}, \ref{fig:pascal_vistas_1},
\ref{fig:city_to_city}, \ref{fig:vistas_to_city}, 
\ref{fig:self_to_self} and \ref{fig:animals_vistas} 
show the responses of OOD detection for various models.
Red denotes the confidence that the corresponding pixel is OOD,
which can also be interpreted as epistemic uncertainty.
The columns in these images correspond to:
    i) original image, 
    ii) max-softmax of the primary model
        (cf.\ Table \ref{table:APseg}), 
    iii) OOD detection with the ROB model
        (cf.\ Table \ref{table:APROB}), 
    iv) discriminative OOD detection trained 
        on entire images from ILSVRC (OOD) 
        and Vistas train (ID)
        (cf.\ Table \ref{table:APbin}), 
    and v) 
      discriminative OOD detection trained 
      on entire ILSVRC images (OOD),
      and ILSVRC bounding boxes (OOD)
      pasted over Vistas images without ground animals (ID),
      as described in Section \ref{ss:traindata}.

These results once again show that the max-softmax approach 
predicts high uncertainty on object borders. 
%It also detects borders of OOD patches 
%but does not have high uncertainty within the patch itself. 
Both the ROB model and the discriminative model trained on entire images 
fail to detect OOD patches in many images
(Figures \ref{fig:pascal_vistas_10}, \ref{fig:pascal_vistas_1} and 
\ref{fig:animals_vistas}). 
Poor performance of the ROB model is expected 
since its training datasets do not include animals.
Poor performance of the discriminative model 
trained on entire images is also understandable
since none of its training images had 
a border between ID and OOD regions.
%The ROB model is a bit more sensitive to Vistas patches in Cityscapes images %(Figure \ref{fig:vistas_to_city}).

The discriminative model which we train 
according to Section \ref{ss:traindata}
delivers the best overall performance. 
It is able to detects OOD patches 
even on very small pasted objects
(cf.\ Figure \ref{fig:pascal_vistas_1})
and genuine animals in Vistas images
(cf.\ Figure \ref{fig:animals_vistas}). 
We note that this model occasionally detects
borders of ID patches (row 2 in Figure \ref{fig:city_to_city} 
and row 2 in Figure \ref{fig:vistas_to_city})
which suggests that results on PascalToVistas
may be a little too optimistic. 
We also note that this model sometimes 
misclassifies parts of Cityscapes images.

Genuine Vistas images with ground animals (VistasAnimals)
is the most difficult dataset for all models,
however the discriminative model 
trained according to Section \ref{ss:traindata} 
clearly achieves the best performance
(cf. Figure \ref{fig:animals_vistas}, row 1).

Table \ref{table:APcontrol} shows average precision 
for pasted content detection on the three control datasets. 
The AP on control datasets indicates 
if the model is able to distinguish 
between the ID image and the ID pasted region. 
High AP on these datasets means that 
the corresponding OOD model 
detects differences in imaging conditions 
or unexpected object locations 
between the ID image and the pasted ID patch.
High performance on control datasets would indicate 
that success on PascalToVistas datasets
is the result of detecting the process of pasting
instead of the novelty of the Pascal classes.
The score of the best discriminative model 
indeed indicates that part 
of its success on PascalToVistas datasets 
comes from recognizing pasting interventions.

\setlength{\tabcolsep}{4pt}
\begin{table}[htb]
\begin{center}
\caption{AP for detection of pasted content 
  in the three control datasets.
  Labels stand for max-softmax of the primary model (ms),
  max-softmax of the model with trained confidence (ms-conf),
  primary model trained for the ROB challenge (ROB),
  and the discriminative model (discrim).
}
\label{table:APcontrol}
\begin{tabular}{|l|l||c|c|c|}
\hline
  Model &
  Training set & 

  CityCity &
  VistasCity &
  Self 

 \\
  \hline
  \hline
  ms &
  city &
  3.96 &
  7.42 &
  5.74 
  
 \\
 \hline
  ms &
  city, wd &
  3.49 &
  6.81 &
  7.42 

 \\
 \hline
  ms-conf &
  city &

  3.61 &
  6.69 &
  5.34 

  \\
  \hline
  ROB &
  ROB &

  4.64 &
  13.37 &
  5.95 

  \\
  \hline
  discrim &
  city, img &

  2.15 &
  45.48 &
  2.92 

  \\
  \hline
  discrim &
  city, wd, img &

  2.68 &
  42.82 &
  3.21 

  \\
  \hline
  discrim &
  vistas, img &

  2.39 &
  9.14 &
  3.56 

  \\
  \hline
  discrim &
  vistas-a, img\_bb &

  7.62 &
  34.12 &
  19.74 
  \\
  \hline
\end{tabular}
\end{center}
\end{table}
\setlength{\tabcolsep}{1.4pt}

\pagebreak
\subsection{Conclusion}

Experiments show that training on ILSVRC bounding boxes 
pasted above Vistas images is able to deliver
fair open-set dense-prediction performance.
In particular, our model succeeds 
to detect animals in road-driving images
although it was not specifically trained for that task,
while outperforming all previous approaches by a wide margin.
We believe that these results strengthen
the conclusions from the main article
and provide useful insights for future work 
in estimating epistemic uncertainty in images.

%in detecting animals in traffic scenes
%with a model that was not specifically trained for animal detection.

\end{appendices}

\end{document}